%% file: main.tex
\documentclass[10pt,twocolumn,letterpaper]{article}

\usepackage[pagenumbers]{wacv} 

\usepackage{graphicx}
\usepackage{amsmath}
\usepackage{amssymb}
\usepackage{booktabs}
\usepackage{algorithm} 
\usepackage{algpseudocode} 
\usepackage[accsupp]{axessibility}
\usepackage{multirow}

\usepackage[dvipsnames]{xcolor}
\usepackage[pagebackref,breaklinks,colorlinks]{hyperref}

\usepackage[capitalize]{cleveref}
\crefname{section}{Sec.}{Secs.}
\Crefname{section}{Section}{Sections}
\Crefname{table}{Table}{Tables}
\crefname{table}{Tab.}{Tabs.}

\def\wacvPaperID{1174} 
\def\confName{WACV}
\def\confYear{2025}

\begin{document}

\definecolor{mygreen}{HTML}{00FF00}

\title{Recurrence-based Vanishing Point Detection}

\author{Skanda Bharadwaj \and Robert T. Collins \and Yanxi Liu \and
School of EECS, 
The Pennsylvania State University, University Park, PA\\
{\tt\small skanda.bharadwaj@psu.edu, \{rcollins, yanxi\}@cse.psu.edu}
}
\maketitle

\begin{abstract}
   Classical approaches to Vanishing Point Detection (VPD) rely solely on the presence of explicit straight lines in images, while recent supervised deep learning approaches need labeled datasets for training. We propose an alternative unsupervised approach: \textit{Recurrence-based Vanishing Point Detection (R-VPD)} that uses implicit lines discovered from recurring correspondences in addition to explicit lines. Furthermore, we contribute two Recurring-Pattern-for-Vanishing-Point (RPVP) datasets: 1) a Synthetic Image dataset with 3,200 ground truth vanishing points and camera parameters, and 2) a Real-World Image dataset with 1,400 human annotated vanishing points.  We compare our method with two classical methods and two state-of-the-art deep learning-based VPD methods. We demonstrate that our unsupervised approach outperforms all the methods on the synthetic images dataset, outperforms the classical methods, and is on par with the supervised learning approaches on real-world images. Code and data can be found here: \href{http://vision.cse.psu.edu/data/data.shtml}{http://vision.cse.psu.edu/data/data.shtml}
\end{abstract}

\section{Introduction}
\label{sec:intro}

\begin{figure}
\centering
\includegraphics[width = \linewidth]{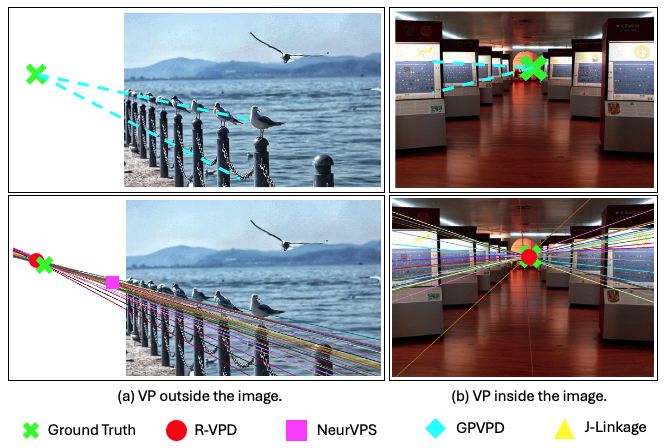}
\caption{Two images containing examples of ``{\em things that recur}" (Recurring Patterns). Top row: images with ground truth VP (\textcolor{green}{$\times$}) indicated. Bottom row: 
VP prediction results from four VPD methods R-VPD (our method),  
NeurVPS \cite{zhou2019neurvps}, 
GPVPD \cite{lin2022deep} and 
J-Linkage \cite{Toldo2008RobustMS} respectively, where  
VP predictions that are too far from the ground truth are not shown in the images ((a) 2 misses, (b) 3 misses).
Due to a lack of explicit straight lines, automatic vanishing point detection on these images poses challenges to explicit line-based methods.} 
\label{fig:rp_ex1}
\vspace{-0.3cm}
\end{figure}

Under perspective projection, parallel lines in 3D meet at a point called the {\it vanishing point} (VP) \cite{mundy1992projective} which may lie inside or outside of the 2D image (Figure \ref{fig:rp_ex1}). Vanishing Point Detection (VPD) plays a vital role in computer vision applications such as structure from motion \cite{gao2017exploiting, li2019line}, 3D reconstruction \cite{guillou2000using, sinha2009piecewise, zhou2019learning}, camera calibration \cite{caprile1990using, cipolla1999camera, coughlan1999manhattan, guillou2000using}, scene understanding \cite{flint2011manhattan}, SLAM \cite{zhou2015structslam, li2018monocular, li2019leveraging}, facade detection \cite{liu2014local} and autonomous driving \cite{lee2017vpgnet}. VPD algorithms can typically be separated into two categories: classical methods that are based on projective geometry \cite{magee1984determining, collins1990vanishing, schaffalitzky2000planar, zhou2017detecting, li2020quasi} and deep learning methods \cite{borji2016vanishing, chang2018deepvp,  zhou2019neurvps, liu2021vapid, lin2022deep}. Classical VPD methods usually consist of two stages - 1) Line Segment Detection (LSD) \cite{denis2008efficient} and 2) VP estimation using line clustering \cite{mclean1995vanishing, bazin2012globally} or voting \cite{gamba1996vanishing, bazin2013globally}. Deep learning VPD methods are data-driven approaches using filters learned by training on hundreds of thousands of annotated images \cite{krizhevsky2012imagenet, karpathy2014large}.

\begin{figure*}[h]
    \centering
    \includegraphics[width = 0.9\linewidth]{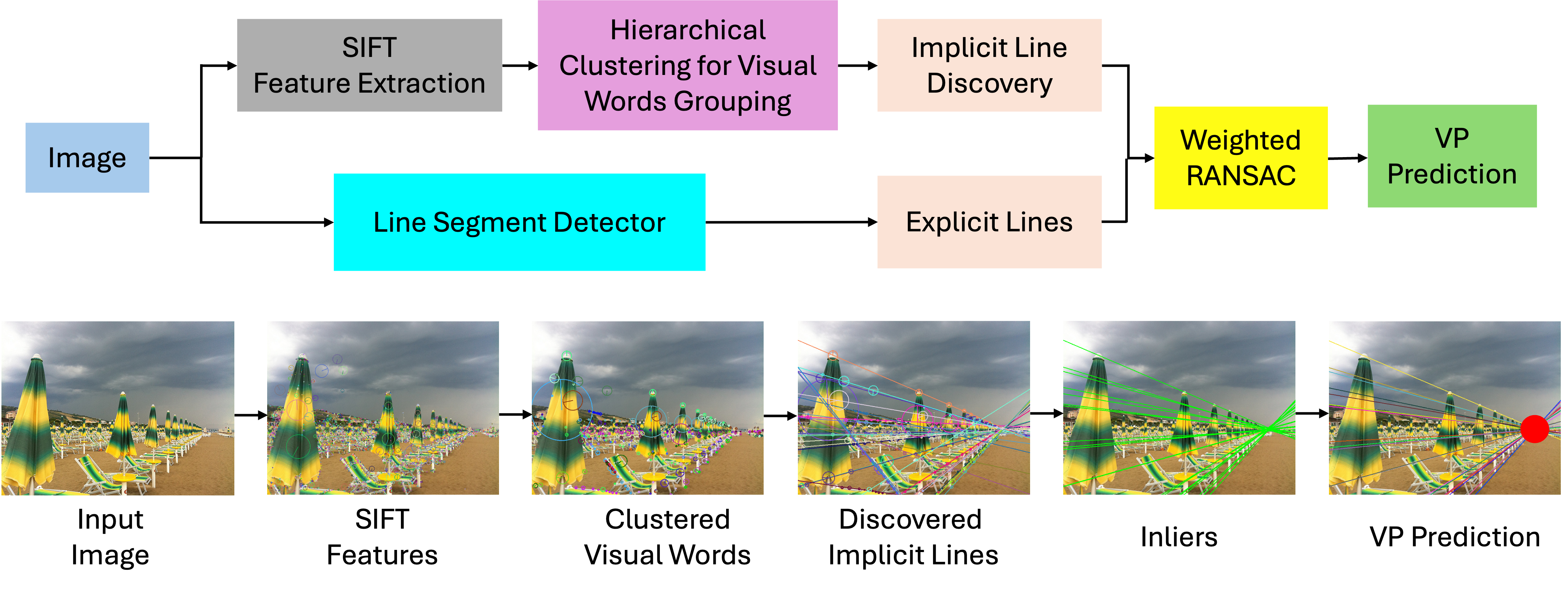}
    \caption{Recurrence-based Vanishing Point Detection: An overview of R-VPD is presented in the above image. SIFT features are extracted from the input image and are clustered hierarchically. Forward selection of the features is performed using geometric constraints based on linearity, angle, and scale. Implicit lines are fitted to the selected feature clusters, and a weighted RANSAC algorithm is employed to find the intersection of both implicit and explicit lines to locate the vanishing point. For this image, there are no explicit long lines.}
    \label{fig:rpvpd}
    \vspace{-0.3cm}
\end{figure*}

We propose an alternative approach to detect VPs using implicit lines discovered from Recurring Patterns (RPs), defined as 
{\em ``things that recur''} \cite{liu2013grasp}, which have common and frequent appearances in real-world images (Figure~\ref{fig:rp_ex1}).
Different from existing methods on VPD, we use groups of corresponding feature points, discovered during the RP detection process,
to construct and validate the co-linearity of implicit lines. We hypothesize intersections of these virtual lines in addition to explicit lines and find the VP through a robust, joint, weighted RANSAC procedure. 
An overview of this approach is shown in Figure~\ref{fig:rpvpd}.
Our Recurrence-based VP method (R-VPD) provides additional robustness for VPD on images without explicit lines leading to VPs.  


\noindent Our contributions include:
\vspace{-0.1cm}
\begin{enumerate}
\setlength{\itemsep}{0.2pt}
\setlength{\parskip}{0pt}
\setlength{\parsep}{0pt}

    \item We propose a novel methodology, \textbf{R}ecurrence-based \textbf{V}anishing \textbf{P}oint \textbf{D}etection (R-VPD), to detect vanishing points in an unsupervised manner using the implicit lines obtained from feature correspondences of RPs in addition to explicit image lines.  

    \item We provide two benchmark datasets - (a) \textbf{Synthetic} images  (RPVP-Synthetic) comprised of 3,200 images created in Blender with the ground truth VP and camera parameters (intrinsic and extrinsic matrices) used to create the 2D images, and (b) \textbf{Real}-world images (RPVP-Real) with 1,400 annotated real-world images containing  RPs, where the ground truth VP is hand-labeled using a {\it LabelMe} tool \cite{LabelMe}.
    
    \item We establish the first VPD benchmark on RP-datasets by comparing four VPD algorithms including two classical approaches: a simple baseline Line Segment Detection (LSD) and J-Linkage \cite{Toldo2008RobustMS} and two state-of-the-art deep learning approaches: NeurVPS \cite{zhou2019neurvps} and GPVPD \cite{lin2022deep}.

\end{enumerate}


\section{Related Work}
\label{sec:rw}
Broadly speaking, VPD methods can be separated into two categories: 1) Classical methods; and 2) Deep learning methods. Early classical methods detect VPs as intersections of straight line segments, typically using the Hough transform \cite{quan1989determining, lutton1994contribution} followed by line clustering \cite{mclean1995vanishing} or other voting schemes \cite{gamba1996vanishing}. 
Barnard \cite{Barnard1983GaussianSphere} introduced a Gaussian sphere representation where lines in the image map to great circles on a sphere, a bounded space more representative of the projective plane than the unbounded image plane and thus better able to support the detection of vanishing points formed by parallel image lines. 
Motivated by this representation, Collins and Weiss   \cite{collins1990vanishing} formulated vanishing point calculation as a statistical estimation problem on the unit sphere, estimating a vanishing point location and associated confidence region as the polar axis of an equatorial distribution on the sphere. Schaffalitzky {\em et al.}~\cite{schaffalitzky2000planar} presented a proof-of-concept planar grouping approach that exploits specific geometric configurations, such as equally spaced coplanar lines, to estimate vanishing points and lines.  Zhou {\em et al.} \cite{zhou2017detecting} addressed the problem of understanding linear perspective in landscape photography, detecting dominant vanishing points by exploiting global structures in the scene via contour detection. One of the most recent VP estimation methods in the ``classical'' vein (Li {\em et al.} \cite{li2020quasi}) makes a Manhattan world assumption to formulate VP estimation as computing the rotation between the Manhattan frame and the camera frame. 

Another category of work leverages groupings of coplanar interest points~\cite{schaffalitzky2000planar,PrittsMinimalSolversPAMI2021,PrittsChumCVPR2014,PrittsEtAlIJCV2020,ChumMatasScaleChange2011} to identify a vanishing line from which the planar surface can be rectified.  In particular, Pritts et.al.~\cite{PrittsMinimalSolversPAMI2021,PrittsChumCVPR2014,PrittsEtAlIJCV2020} leverage differing sizes of conjugately translated planar elements to estimate the vanishing line, allowing affine rectification, and with pattern-dependent constraints sometimes improving the rectification to a similarity transformation. Unlike most other works in projective geometry, they are able to estimate and undo the nonlinear warping effects of radial lens distortion. 

With recent advancements in deep learning, most research focuses on solving problems using learnable filters \cite{chang2018deepvp, zhou2019neurvps, lin2022deep}. NeurVPS \cite{zhou2019neurvps} introduced an efficient VP estimation method using a conic convolution operator that extracts and aggregates features along structural lines. Building on NeurVPS and the early work of Barnard~\cite{Barnard1983GaussianSphere}, Lin {\em et al.}~\cite{lin2022deep} propose two trainable geometric priors, the Hough transform and Gaussian sphere representation, to reduce network sensitivity to dataset variations. Their method first extracts features with a stacked hourglass network \cite{newell2016stacked}, maps them to a Hough space to identify lines, and then uses spherical convolution on the Gaussian sphere to detect VPs.  A recent work  \cite{tong2022transformer} by Tong {\em et al.} does not have publicly available code, making direct comparison infeasible.

Most classical and deep-learning VPD methods depend on explicit line segments in the image. Even the state-of-the-art \cite{zhou2019neurvps} relies on a sufficient number of edges oriented towards the VP. To address VP estimation in images lacking explicit line segments, we propose a novel method that leverages recurring patterns and their properties, which offer strong cues about the 3D world from a 2D image.


\section{R-VPD: Recurrence-based Vanishing Point Detection}
\label{sec:VP_algo}

Recurring Patterns (RPs) are a set of {\em ``things that recur''} 
\cite{liu2013grasp}. 
RPs embody the powerful organizing principle that things that 
co-occur are not accidental.

\begin{figure}[h!]
    \includegraphics[width=0.9\linewidth]{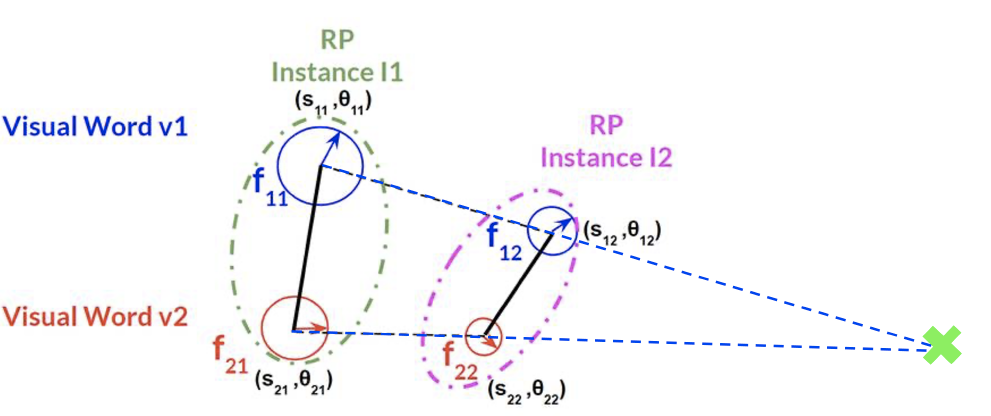}
    \caption{ 
     A Unit Recurring Pattern (URP) is an RP with at least two RP instances (RPIs) and each RPI has at least two distinct visual words (e.g.~clustered SIFT descriptors) that correspond across the RPIs (this figure is adapted from \cite{zhang2022novel} ).  
    Here we add the two blue dotted lines passing through pairs of corresponding features, $(f_{11},f_{12})$ and $(f_{21},f_{22})$ , whose intersections identify the VP.
    } 
    \label{fig:grasp}
    \vspace{-0.3cm}
\end{figure}

A {\em unit RP} (URP), namely the smallest RP 
(Figure~\ref{fig:grasp}) is composed of two {\em RP Instances} (RPI), and an RPI has to contain at least one pair of distinct visual words. We use the SIFT feature extractor to generate the pool of candidate features for forming visual words in an image. The initial phase of SIFT involves constructing a difference of Gaussian (DoG) image pyramid.  Feature points are identified by searching for local extrema across different scales and spatial positions. Once features are localized, a 128-dimensional custom vector is calculated as the feature descriptor. The 64x64 regions surrounding the key points are subdivided into 16 smaller blocks of size 4x4, and an eight-bin orientation histogram is created, resulting in the final generation of a 128-dimensional vector.

Starting with a URP, a 2x2 matrix is constructed with each row being a visual word and each column being an RP instance. An unsupervised, randomized optimization algorithm called GRASP \cite{liu2013grasp} performs a set of random ``moves''  that progressively grow, shrink, or replace entries in the matrix, seeking to maximize its size while being constrained by a set of numeric consistency measures. For more information on optimization-based RP detection, we refer readers to \cite{liu2013grasp}. 
The authors of \cite{zhang2022novel} 
develop an alternative two-stage unsupervised architecture for RP discovery instead of GRASP, yielding enhanced speed.

%

%
%

\subsection{Implicit Recurring Correspondences Discovery}

Different from both \cite{liu2013grasp, zhang2022novel}, we do not explicitly generate RPs from a given image but employ in novel ways the corresponding visual words discovered during the optimization process to fit and merge implicit lines with explicit lines for subsequent VP detection. 
By fitting lines through two or more such correspondences and finding their intersections we can estimate a potential VP. 
Hence we give the name of our approach {\it Recurrence}-based VPD.

The basic difference between our method and existing VPD algorithms is the  use of implicit lines, such that our method does not rely only on explicit edge in an image. Traditional VP detection methods fail in scenes where there are not enough salient line segments in the image (Figure~\ref{fig:rp_ex1}), while our method takes advantage of implicit lines together with any explicit lines to locate the VP. 

\subsection{Hierarchical clustering}
\label{sec:hc}
We use hierarchical clustering, a traditional unsupervised method that groups high-dimensional features into a clustering tree, to further group visual words for line fitting. Unlike other methods, it does not require a preset number of classes, making it ideal for this scenario. A feature-matching matrix is first created using Euclidean distances between features. Hierarchical clustering is then performed by iteratively merging the closest subgroups of features until all features form a single group.

Specifically, let $S$ be the set of feature groups and let $A, B \in S$. Let $N_A$ and $N_B$ be the number of features within the groups $A$ and $B$, respectively. The distance between the groups $A$ and $B$ is defined as:
\begin{equation}
\begin{split}
Dist_{AB} &= min\{dist_{ij}\} , \forall i \in [1, N_A], \forall j \in [1, N_B] \\
dist_{ij} &= {\lVert fd_i - fd_j \rVert}_2
\end{split}
\end{equation}
where $dist_{ij}$ represents the distance between features $i$ and $j$ and $fd_k$ represents the 128-dimensional descriptor of feature $k$. Then, merging the closest neighboring groups is performed simply as  $C = (A \cup B)$, and distances are updated between this new group and existing groups using the equation:
\begin{equation}
    Dist_{CP} = min\{Dist_{AP}, Dist_{BP}\}, \forall P \in \{S \setminus \{A, B\}\} \;.
\end{equation}

\begin{figure}[h!]
    \includegraphics[width=0.9\linewidth]{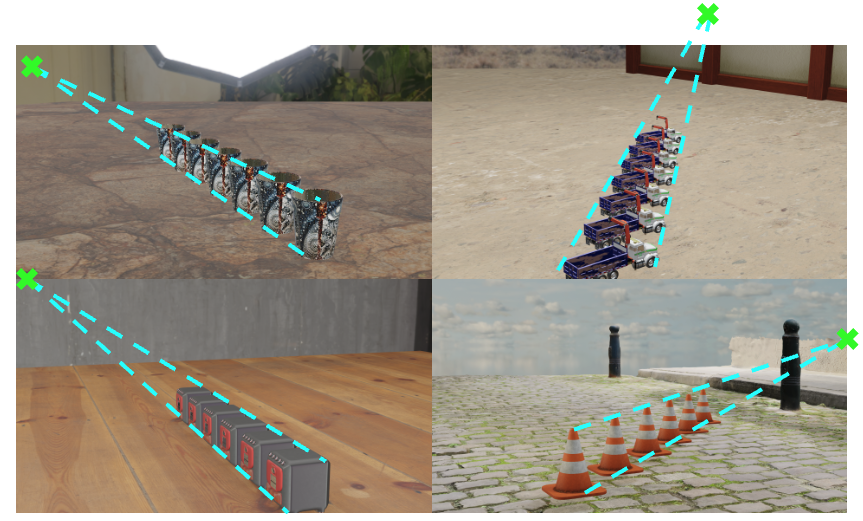}
    \caption{Sample images from RPVP-Synthetic image dataset showing scenes in which RPs are oriented towards the VP. A relatively lesser number of explicit lines makes this dataset challenging for VPD. The "\textcolor{green}{$\times$}" represents the ground truth VP.} 
    \label{fig:rpvp_synth_ex}
    \vspace{-0.3cm}
\end{figure}

\subsection{Forward Feature Selection using Bottom-up Search}
\label{sec:cr_search}

Forward feature selection identifies features within a visual word that are best suited for fitting implicit lines. This is achieved using three  scores—Linearity, Angle, and Scale—that assess distinct aspects of geometric consistency:

\textbf{Linearity Score ($S_L$):} This score measures how well the $(x, y)$ locations of keypoints align along a dominant orientation. It is calculated as the average perpendicular distance of keypoints from a fitted line. A lower score indicates stronger alignment, making the features more suitable for line fitting.

\textbf{Angle Score ($S_A$):} This evaluates the consistency of directional changes across an ordered sequence of keypoints. For triplets $(A, B, C)$, it calculates the absolute difference in angular changes between segments $A \to B$ and $B \to C$. These differences are averaged across all triplets, with lower scores reflecting greater uniformity in direction.

\textbf{Scale Score ($S_S$):} This assesses the consistency of relative size changes between keypoints, influenced by projective transformations. For triplets $(A, B, C)$, it calculates the absolute difference in scale ratios between $A \to B$ and $B \to C$. These differences are averaged, with lower scores indicating more uniform scale progression.

\textbf{Composite Score ($S_C$):} The overall suitability of a feature set is determined by: $S_C = S_L \times \exp(S_A + S_S) / N^2$, where $N$ is the number of features in the visual word. This score integrates collinearity, angular consistency, and scale variability while normalizing for feature set size, ensuring that only geometrically coherent features are selected.

By combining these metrics, forward feature selection prioritizes features that exhibit strong geometric consistency, enhancing the reliability of vanishing point estimation. A more elaborate explanation of the scores can be found in the Appendix.

\subsection{Implicit Line Fitting and Weighted RANSAC}
\label{sec:linefit}
Given a set of corresponding SIFT features, we can fit an orientation vector using the least squares method. The orientation of this vector is determined by the scale of the associated features, with vectors from larger to smaller scale features regarded as pointing toward the vanishing point. A weighted RANSAC algorithm is employed to remove outliers. This algorithm entails randomly sampling two lines based on their weights (see below), calculating their intersection, and tallying the votes for this particular prediction among all other lines. These steps are repeated iteratively until the optimal prediction with the highest number of votes is identified. The vanishing point location is ultimately estimated by least squares from all the inliers.


The initial weight $w_i$ for each line $l_i$ is computed by evaluating the angular relationships between $l_i$ and every other line $l_j$ in the dataset. Specifically, the weight $w_i$ is determined 
as: $w_i = \sum_{\substack{j=1 \\ j \neq i}}^{n} e^{-\theta_{ij}}$, where $\theta_{ij}$ is the acute angle between lines $l_i$ and $l_j$. 
This acute angle is calculated as the arccos of the dot product of the direction vectors of the two lines, and if this is an obtuse angle it is subtracted from $\pi$ to ensure that $0 \leq \theta_{ij} \leq \pi/2$. 
The exponential decay function emphasizes smaller angles, thereby assigning higher weights to lines that are closer to being parallel, reflecting a higher probability of contributing to the detection of a consistent vanishing point.

During each iteration of the Weighted RANSAC algorithm, an intersection is estimated and lines are classified as inliers based on their distance to the point. The weights are then updated to refine the selection process in subsequent iterations. For inliers $(l_i \in I)$, the weight $w_i$ is incremented as 
$w'_i = w_i \cdot \alpha$ and for outliers $(l_j \in O)$, the weight $w_j$ is decremented as $w'_j = w_j \cdot \beta$, where $\alpha = 1.2$ and $\beta = 0.8$ in our algorithm.

We note that the intersection of near-parallel lines is ill-conditioned and the intersection points could be incorrect. To remedy this, we run the weighted RANSAC multiple times by reinitializing the weights and adaptively finding an optimal threshold to find inliers. Finally, the vanishing point associated with the largest set of inliers is calculated using eigenvalue decomposition.

\begin{table*}[h!] 
\centering
\caption{RPVP-Synthetic dataset (3200 images) performance measured by $AA^{\theta^\circ}$ representing the {\em AUC} of {\em angle accuracy} at the threshold angle $\theta^\circ$. The {\it p}-values are computed against R-VPD (Ours) to indicate whether the difference in performance is statistically significant. Corresponding \textit{AA} curves are shown in Figure~\ref{fig:sr_rpvp_si}.}
\label{tab:sr_rpvp_si}
\vspace{-0.1cm}
\begin{tabular}{p{3.2cm}|c c c c c|}
    Method  &  $AA^{2^\circ}(\uparrow)$ &  $AA^{5^\circ}(\uparrow)$ &  $AA^{10^\circ}(\uparrow)$ & median$(\downarrow)$  & {\em p}-value 
    \\ \hline 
    LSD                                     &  0.14 & 0.45 & 1.12 & 45.47$^{\circ}$  &  1.80e-16   \\
    J-Linkage \cite{Toldo2008RobustMS}      &  0.42 & 1.46 & 3.43 & 29.41$^{\circ}$  &  5.78e-12  \\
    GPVPD-nyu \cite{lin2022deep}            &  0.04 & 0.26 & 0.71 & 88.38$^{\circ}$  &  1.73e-10   \\ 
    GPVPD-su3 \cite{lin2022deep}            & 0.002 & 0.02 & 0.15 & 73.88$^{\circ}$  &  1.03e-18   \\
    GPVPD-scannet \cite{lin2022deep}        & 0.004 & 0.04 & 0.22 & 71.01$^{\circ}$  &  1.56e-18   \\
    NeurVPS-su3 \cite{zhou2019neurvps}      &  0.17 & 0.72 & 1.88 & 91.83$^{\circ}$  &  9.37e-17   \\
    NeurVPS-scannet \cite{zhou2019neurvps}  &  0.19 & 0.74 & 1.80 & 96.94$^{\circ}$  &  3.06e-16   \\
    NeurVPS-tmm17 \cite{zhou2019neurvps}    &  0.32 & 1.26 & 3.28 & 12.77$^{\circ}$  &  1.15e-07   \\
    \textbf{R-VPD}                           &  \textbf{0.97} & \textbf{3.49} & \textbf{7.44} & \textbf{0.60$^{\circ}$ } &    -     \\
\end{tabular}
\end{table*}

\begin{table*}[h!] 
\centering
\caption{RPVP-Real dataset (1400 images) performance measured by $AA^{\theta^\circ}$ and {\em p-value}. Corresponding \textit{AA} curves are shown in Figure~\ref{fig:sr_rpvp_ri}.\hfill\ }
\label{tab:sr_rpvp_ri}
\vspace{-0.1cm}
\begin{tabular}{p{3.2cm}|c c c c c|}
    Method  &  $AA^{2^\circ}(\uparrow)$ &  $AA^{5^\circ}(\uparrow)$ &  $AA^{10^\circ}(\uparrow)$ & median$(\downarrow)$  & {\em p}-value 
    \\ \hline 
    LSD                                    &  0.36 & 1.29 & 2.96 & 42.95$^{\circ}$  &   4.37e-13   \\
    J-Linkage \cite{Toldo2008RobustMS}     &  0.68 & 2.58 & 6.10 & 1.47$^{\circ}$  &   5.27e-04   \\
    GPVPD-nyu \cite{lin2022deep}           &  0.19 & 1.05 & 3.06 & 71.53$^{\circ}$  &   3.12e-11   \\ 
    GPVPD-su3 \cite{lin2022deep}           &  0.05 & 0.36 & 1.38 & 22.04$^{\circ}$  &   4.82e-14   \\
    GPVPD-scannet \cite{lin2022deep}       &  0.04 & 0.31 & 1.19 & 26.75$^{\circ}$  &   6.42e-15   \\
    NeurVPS-su3 \cite{zhou2019neurvps}     &  0.04 & 0.17 & 0.52 & 88.70$^{\circ}$  &  6.14e-18   \\
    NeurVPS-scannet \cite{zhou2019neurvps} &  0.23 & 1.12 & 2.92 & 32.35$^{\circ}$  &  3.71e-12   \\
    \textbf{R-VPD}                          &  \textbf{1.03} & \textbf{3.75} & \textbf{8.63} & \textbf{0.74$^{\circ}$}  &    -     \\
\end{tabular}
\end{table*}

\section{Experimental Setup}
\label{sec:exp}

\subsection{Datasets}
\label{supsec:dataset}

We evaluate and compare several VPD algorithms on four different datasets. Two of the datasets, RPVP-Synthetic and RPVP-Real, are created/collected and labeled by us. They have  the property that each image contains one or more recurring patterns (RPs) but not necessarily  explicit straight lines.  
We also consider the existing TMM17 dataset used in \cite{zhou2017detecting,zhou2019neurvps} containing natural scenes with one dominant vanishing point, some featuring long, straight lines leading to the VP but not necessarily recurring patterns.  Specifically, we use the test-image set used by \cite{zhou2019neurvps}, TMM17-Test, to evaluate all the VPD methods compared in this paper.  
Furthermore, we note there is some overlap between images containing RPs in our RPVP-Real and TMM17. Since NeurVPS-tmm17 \cite{zhou2019neurvps} is trained on a large portion of TMM17, to avoid data leakage when comparing against that algorithm on RPVP-Real we form a subset RPVP-Real-Exclusive of images in RPVP-Real that are not found in TMM17.

\subsubsection{RPVP-Synthetic: Recurring Pattern-based Vanishing Point Dataset with Synthetic Images} 
A custom dataset is created using \textit{Blender}, an open-source 3D computer graphics software tool \cite{Blender}.  We generate 3,200 images from 16 different synthetic 3D scenes (shown in Figure \ref{fig:rpvp_synth_ex}), each contributing 200 images. In this dataset, objects are placed along a straight line and oriented toward a vanishing point, with each scene featuring varying objects, backgrounds, scales, and camera orientations. Ground truth annotations including the $(x, y)$ coordinates of the vanishing point and the camera parameters are provided for each image. 
More information on scene creation and frame extraction is provided in the Appendix. It is worth noting that a majority of images from this dataset do not contain explicit straight lines oriented towards the vanishing point.  

\subsubsection{RPVP-Real: Recurring Pattern-based Vanishing Point Dataset with Real-world Images} 
\label{sec_RPVP-Real}

RPVP-Real contains 1400 ``real world" images with VPs that are produced from recurring patterns under perspective projection.  Explicit straight lines may or may not be available. 
The VPs on these images are manually labeled using the \textit{LabelMe} tool \cite{LabelMe} and ground truth annotations are provided as text files for each image with the $(x, y)$ coordinates of the vanishing point.  
The ground truth vanishing point is then calculated as the intersection of these annotated lines.

\subsubsection{TMM17-Test: Natural Scenes Dataset}
\label{sec:tmm17}
The TMM17 dataset \cite{zhou2017detecting,zhou2019neurvps} consists of 2275 images of natural scenes with one dominant vanishing point. As illustrated in Figure \ref{fig:dataset} 
TMM17 overlaps with over 70\% of RPVP-Real images (images with recurring patterns).
%
The one split\cite{zhou2019neurvps} used 2000 images for training and 275 images as the test image set. 
To avoid any data leaking, we choose TMM17-Test, as a validation image set for all VPD methods. 
%
It is worth noting that TMM17-Test contains both images with and without RPs (Figure \ref{fig:dataset}).
Even though R-VPD is designed primarily to deal with images with RPs, we use TMM17-Test to test the robustness of R-VPD on real world scenes beyond those with recurring patterns. 

\subsubsection{RPVP-Real-Exclusive}

\begin{figure}[h!]
\centering
\includegraphics[width=0.7\linewidth]{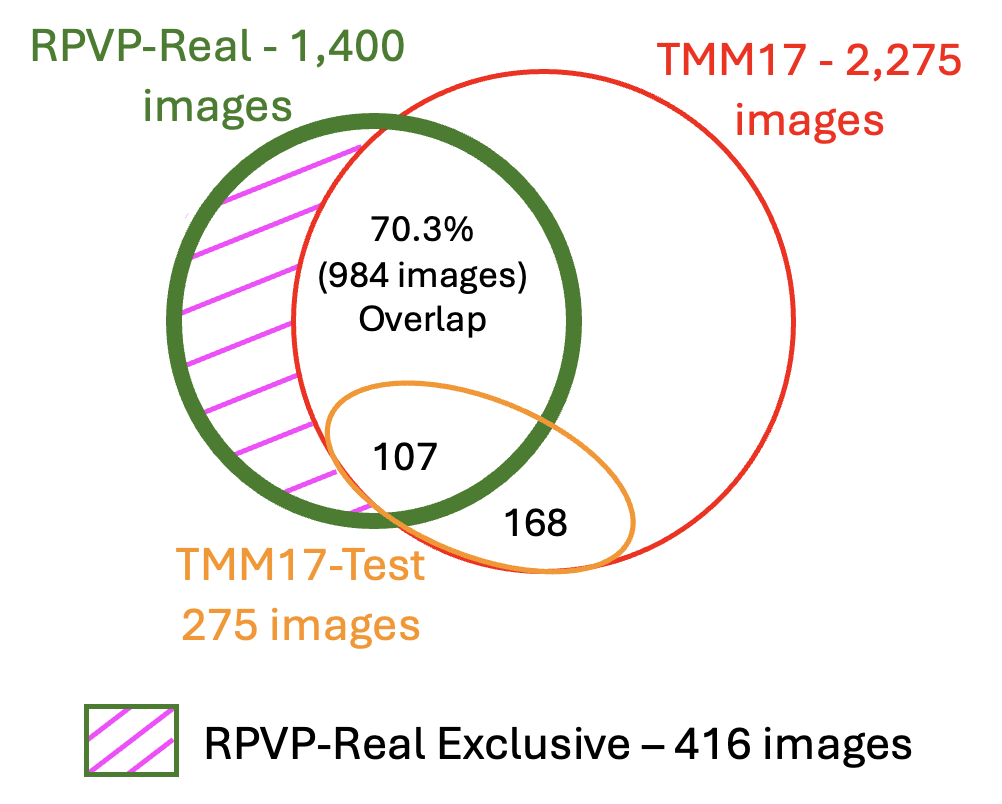}
\vspace{-0.2cm}
\caption{The relationship between RPVP-Real, TMM17, RPVP-Real Exclusive and TMM17-Test.  TMM17/TMM17-Test contains natural images with/without RPs while RPVP-Real/RPVP-Real Exclusive contains images with RPs only.}
\label{fig:dataset}
\vspace{-0.3cm}
\end{figure}

The RPVP-Real dataset has an overlap of about 70.3\% with the TMM17 dataset (Sec \ref{sec:tmm17}) used to train NeurVPS-tmm17 \cite{zhou2019neurvps}. To avoid potential data leaking when comparing with NeurVPS, we form an additional dataset which is the subset (29.7\%) of RPVP-Real that has no overlap with TMM17, which we call RPVP-Real Exclusive dataset that consists of 416 images (Figure~\ref{fig:dataset}). 

\subsection{Evaluation Metrics}
\label{sec:eval_met}
We compare our VP detection method with two classical methods: a baseline Line Segment Detection (LSD) method and a more sophisticated J-Linkage \cite{Toldo2008RobustMS} approach.  We also compare two state-of-the-art deep learning methods, NeurVPS \cite{zhou2019neurvps} and GPVPD \cite{lin2022deep}. For NeurVPS, we consider three networks: 1) NeurVPS-tmm17, 2) NeurVPS-su3 and 3) NeurVPS-scannet, where NeurVPS is trained on TMM17 \cite{zhou2017detecting}, SU3 \cite{zhou2019learning}, or ScanNet \cite{dai2017scannet} dataset, respectively. Similarly, for GPVPD, we compare with three variants of their model since they claim in \cite{lin2022deep} that geometric priors make the dataset variations vanish. The three variants are 1) GPVPD-nyu, 2) GPVPD-su3, and 3) GPVPD-scannet, where GPVPD is trained on NYU \cite{silberman2012indoor}, SU3 \cite{zhou2019learning}, or ScanNet \cite{dai2017scannet} dataset, respectively. With our RPVP datasets we emphasize different algorithms failing to find a VP in the absence of explicit line segments.

\begin{figure}[h!]
\centering
\includegraphics[width=0.7\linewidth]{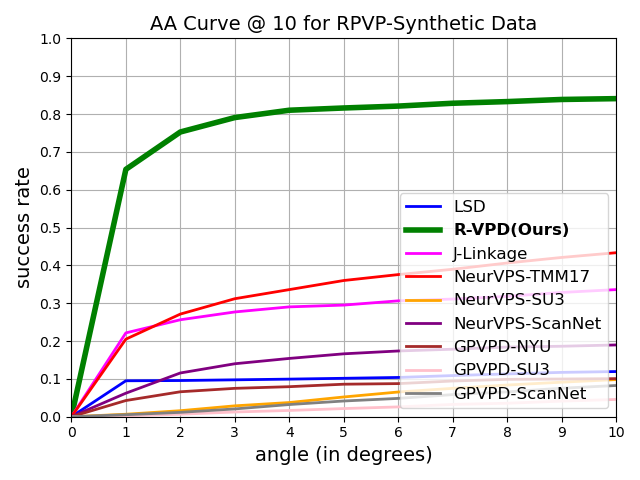}
\vspace{-0.2cm}
\caption{RPVP-Synthetic dataset (3200 images): Success rate curves for {\it AA @ $10^\circ$}. 
Corresponding \textit{AUC} is given in Table~\ref{tab:sr_rpvp_si}.}
\label{fig:sr_rpvp_si}
\vspace{-0.3cm}
\end{figure}

\begin{figure}[h!]
\centering
\includegraphics[width=0.7\linewidth]{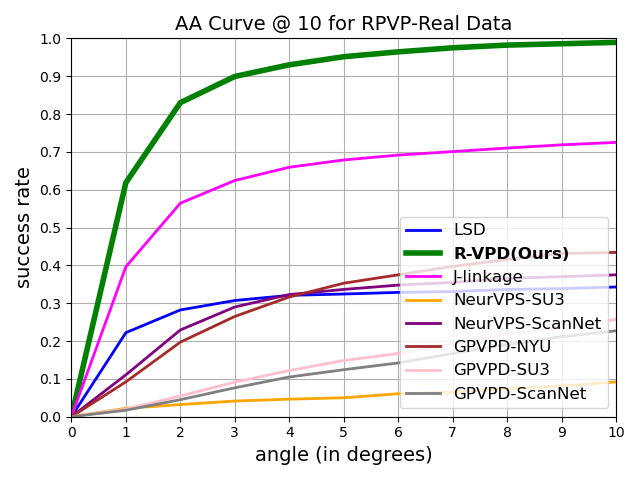}
\vspace{-0.2cm}
\caption{RPVP-Real dataset (1400 images): Success rate curves for {\it AA @ $10^\circ$}. Corresponding \textit{AUC} and are given in Table~\ref{tab:sr_rpvp_ri}.}
\label{fig:sr_rpvp_ri}
\vspace{-0.3cm}
\end{figure}

We use a vector orientation-based method for comparing a detected VP and the ground truth. In this method, the VP is represented as a unit vector in the direction $(\mbox{vp}_x - x_0, \mbox{vp}_y - y_0, f)$, where $(x_0,y_0)$ is the image center, and $f$ is a nominal value that would be the camera focal length if known, but otherwise is chosen to be (image width + image height)/4. Distance in this case is the angle between the detected and ground truth VP unit vectors. We call this metric {\em angle accuracy} ({\em AA}) (also used in \cite{zhou2019neurvps}). With this metric, we can plot the {\em success rate curves} for {\em AA} for each method by varying the angle thresholds. The success rate of VP detection is computed as the ratio of the number of acceptable VPs against ground truth over all detected VPs. 

Along with success rate curves we also provide the {\em AUC} (area under the curve) to evaluate the performance of different algorithms. For {\em AUC} we evaluate all algorithms at thresholds $AA = [2, 5, 10]^{\circ}$. In our {\em AUC} tables, $AA^{\theta^\circ}$ represents the {\em AUC} of {\em angle accuracy} at threshold angle $\theta^\circ$.

\section{Results and Discussions}
\label{sec:res}

\subsection{P-values}
Table~\ref{tab:sr_rpvp_si} provides {\em AUC} (Figure \ref{fig:sr_rpvp_si}) for all the methods on the RPVP-Synthetic dataset at thresholds $AA = [2, 5, 10]^{\circ}$ respectively. The last column of the table reports \textit{p}-values, a measure that indicates whether a difference in numerical performance between our method and each of the other methods is statistically significant, and is based on comparing the entire {\it AUC} curve (all the possible thresholds) between our method and the other methods. 
Although not commonly used in computer vision (but widely used in scientific domains such as biomedical research), {\it p-values} indicate whether the numerical results of one method are {\em statistically significantly different} from another, that is, whether the difference in performance is likely due to a real effect or just random chance. Typically one chooses a threshold of 0.05 or 0.001 to classify {\it p-values} into significant (lower than threshold) or not significant (higher than threshold). 

Table~\ref{tab:sr_rpvp_ri} and Figure \ref{fig:sr_rpvp_ri} provide results on the RPVP-Real dataset. 
NeurVPS-tmm17 is excluded from this comparison due to the overlap between RPVP-Real and TMM17, however a direct and fair comparison between our method and and NeurVPS-tmm17 will be presented below.

Both quantitative results on RPVP-Synthetic and RPVP-Real show that our method R-VPD outperforms the other compared methods {\em significantly} ({\em p-value} $<<$ 0.001). This suggests that classical and current learning-based VPD methods struggle when explicit straight-line segments are absent. 

\subsection{RPVP-Real-Exclusive: RP Image Exclusive}

\begin{figure}[h!]
\centering
\includegraphics[width=0.7\linewidth]{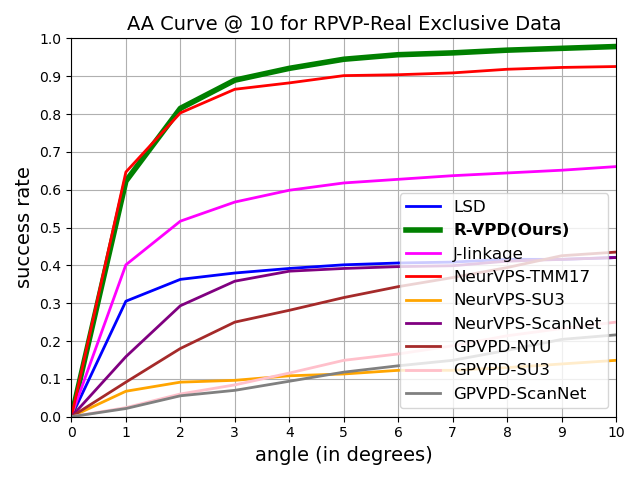}
\vspace{-0.2cm}
\caption{Success Rate curves for the different methods on RPVP-Real Exclusive dataset (416 images). 
Corresponding {\em AUC} is given in Table \ref{tab:sr_rpvp_ri_ex}.}
\label{fig:sr_rpvp_ri_ex}
\vspace{-0.4cm}
\end{figure}

To avoid any potential data leaking issues, we restrict our evaluation of NeurVPS-tmm17 to the RPVP-Real Exclusive dataset that is non-overlapping with TMM17 (Figure \ref{fig:dataset}). 
Here, our R-VPD method performs slightly better than NeurVPS-tmm17 (Table~\ref{tab:sr_rpvp_ri_ex}, Figure~\ref{fig:sr_rpvp_ri_ex}) with {\em p-value} = 0.58. Thus, we can state that our method is on par with the state of the art deep learning method such as NeurVPS-tmm17 on VPD.




\begin{figure}[h!]
\centering
\includegraphics[width=0.7\linewidth]{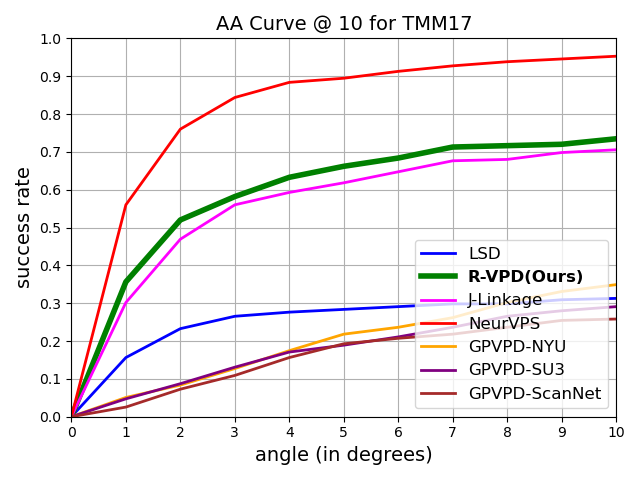}
\vspace{-0.2cm}
\caption{Success rate curves for {\it AA @ $10^\circ$} for TMM17-Test set (275 images). Corresponding \textit{AUC} is given in Table~\ref{tab:sr_tmm17}.}
\label{fig:sr_tmm17}
\vspace{-0.2cm}
\end{figure}

\begin{table}[h!] 
\centering
\caption{Performance of NeurVPS and R-VPD on RPVP-Real Exclusive dataset (416 images). Corresponding \textit{AA} curves are shown in Figure \ref{fig:sr_rpvp_ri_ex}.}
\vspace{-0.2cm}
\label{tab:sr_rpvp_ri_ex}
\small
\setlength{\tabcolsep}{3pt} 
\begin{tabular}{p{2.1cm}|cccccc|}
    Method  &  $AA^{5^\circ}(\uparrow)$ &  $AA^{10^\circ}(\uparrow)$ & median $(\downarrow)$ & {\em p}-value
    \\ \hline 
    NeurVPS \cite{zhou2019neurvps}    & 3.64 & 8.21 & 0.72$^{\circ}$ & 0.58\\
    \textbf{R-VPD(Ours)}                     & \textbf{3.71} & \textbf{8.54} & \textbf{0.71$^{\circ}$}  & -\\
\end{tabular}
\vspace{-0.2cm}
\end{table}

\begin{table}[tb!] 
\centering
\caption{Performance of all VPD methods on TMM17-Test set (275 images, Fig. \ref{fig:dataset}). Corresponding \textit{AA} curves are shown in Fig~\ref{fig:sr_tmm17}.}
\label{tab:sr_tmm17}
\vspace{-0.1cm}
\small
\setlength{\tabcolsep}{3pt} 
\begin{tabular}{p{2.2cm}|c c c c c c|}
    Method  &  $AA^{5^\circ}(\uparrow)$ &  $AA^{10^\circ}(\uparrow)$ & median $(\downarrow)$ & {\em p}-value \\ \hline 
    LSD                                    & 1.07  & 2.56 & 71.30$^{\circ}$ &  8.41e-10\\
    J-Linkage \cite{Toldo2008RobustMS}     & 2.23 & 5.59 & 2.42$^{\circ}$ & 0.52\\
    \small GPVPD-su3 \cite{lin2022deep}        & 0.53 & 1.76 & 31.83$^{\circ}$ & 8.16e-10\\ 
    \small GPVPD-nyu            & 0.55 & 1.96 & 52.80$^{\circ}$ & 7.43e-09\\ 
    \small GPVPD-scannet      & 0.45 & 1.60 & 35.05$^{\circ}$ & 6.81e-11\\ 
    NeurVPS \cite{zhou2019neurvps}    & \textbf{3.49} & \textbf{8.14} & \textbf{0.86$^{\circ}$} & 0.003\\
    \textbf{R-VPD(Ours)}                      & 2.42 & 5.95 & 1.85$^{\circ}$ & - \\
\end{tabular}
\vspace{-0.2cm}
\end{table}

\begin{figure*}[h!]
\centering
\includegraphics[width = 0.9\linewidth]{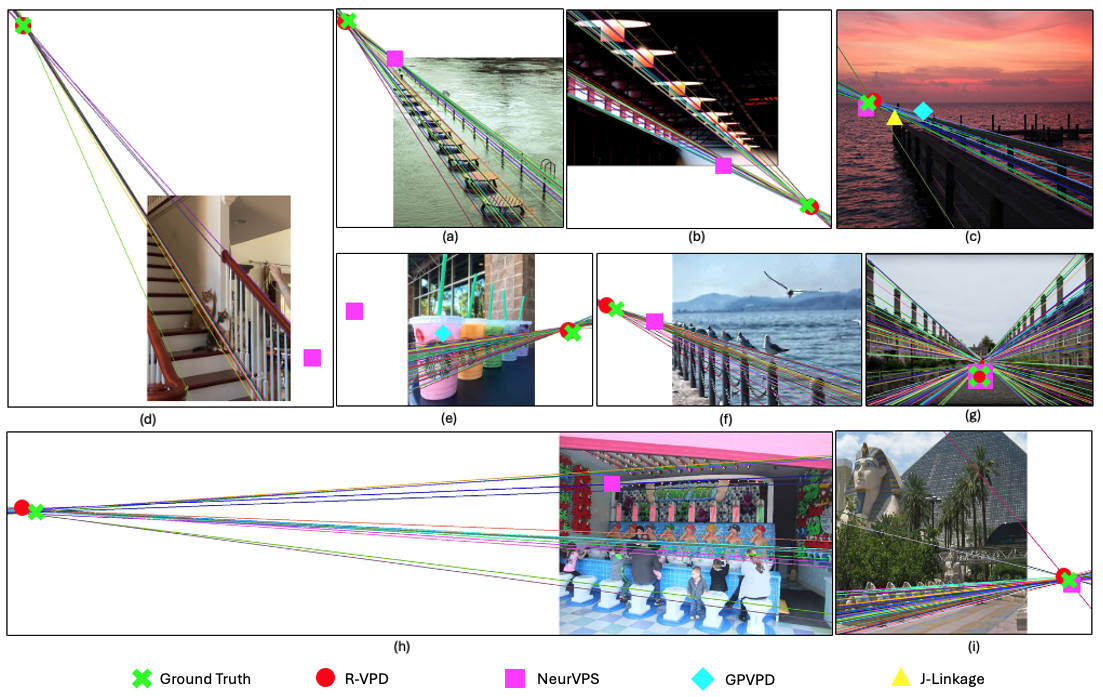}
\caption{Sample results for VP detection on real-world images using our method and a comparison with other approaches. The absence of a specific symbol in any image indicates the failure of the corresponding method to detect the vanishing point in that image.}
\label{fig:vp_ex_1}
\vspace{-0.3cm}
\end{figure*}

\begin{figure}
\centering
\includegraphics[width=0.7\linewidth]{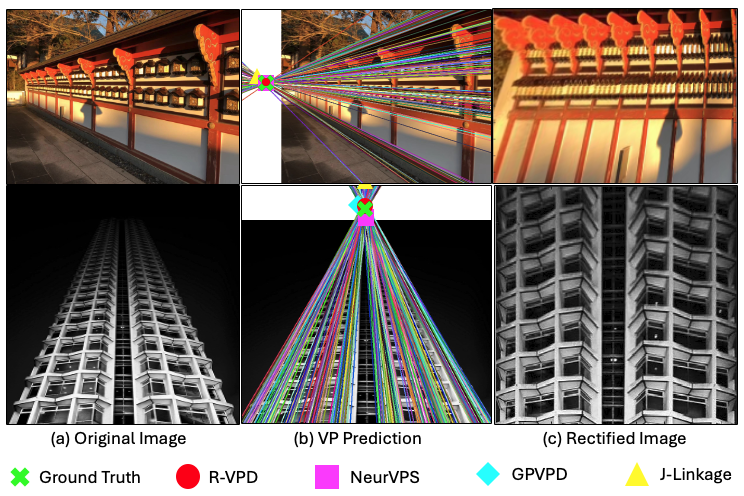}
\caption{One application using a single discovered VP in one-point perspective images is to perform  rectification  of the original image into an affine (parallelism preserving) frontal view.}
\label{fig:rp_rectify}
\end{figure}

\subsection{TMM17-Test: Going beyond RP images}
\label{sec:TMM17-Test}

To evaluate on images that do not necessarily have recurring patterns in them, we perform an evaluation on the same test set used in \cite{zhou2019neurvps}, TMM17-Test.
Figure~\ref{fig:sr_tmm17} and Table~\ref{tab:sr_tmm17} provide {\em AUC} for all the methods on TMM17-Test. R-VPD performs second to the best on TMM17-Test without any training, and with a p-value of $p=0.003$ ($>  0.001$) the best performer NeurVPS-tmm17 does not significantly outperform our approach. 
It is worth noting that of the 275 images in the TMM17-Test set, a considerable portion has no RP on the images (Figure \ref{fig:dataset}).


\subsection{Qualitative Evaluation and Speed}

Figures~\ref{fig:vp_ex_1} and \ref{fig:rp_rectify} present results on some representative images, showing the effectiveness of 
our R-VPD method compared with ground truth and all the SOTA methods. 
Classical LSD-based algorithms 
depend on explicit line segments in images \cite{zhang2022novel} while 
learning-based approaches are  learning edge filters oriented locally towards the VP.
For this reason, real-world images that do not have explicit straight lines or edges oriented towards the VP can lead these VPD methods to make erroneous predictions of the VP (Figure \ref{fig:vp_ex_1}).


Processing time comparisons are shown in Table~\ref{tab:processing_time}. 
NeurVPS and GPVPD achieve their speed by using a GPU and require up to 400 MB of memory, which can be limiting in memory-constrained settings. J-Link appears to be faster here due to fewer lines to process. 

The complexity of our algorithm depends on the number {\em n} of SIFT features detected and is dominated by the calculation of the distance matrix between these features. Therefore our algorithm's time and space complexity is $O(n^2)$.

\begin{table}[h!]
\centering
\small
\caption{Processing time per image comparison (median times).\hfill\ }
\vspace{-0.2cm}
\label{tab:processing_time}
\begin{tabular}{c|ccc}
Method  & Hardware & RPVP-Real & RPVP-Synth \\ 
\hline
NeurVPS   & \multirow{2}{*}{GPU} & 0.86s  & 0.52s \\ 
GPVPD     &                      & 0.52s  & 0.52s  \\ 
J-Link    & \multirow{2}{*}{CPU} & 18.67s  & 0.75s  \\
R-VPD     &                      & 10.73s  & 1.31s \\    
\end{tabular}
\vspace{-0.3cm}
\end{table}

\section{Conclusion}
We have proposed an alternative unsupervised vanishing point detection algorithm, R-VPD, that uses feature correspondences in recurring patterns (RP) to form implicit lines for VPD. We contribute two RP-based vanishing point datasets: RPVP-Synthetic has 3200 synthetic images with VPs and camera parameters, and RPVP-Real has 1400 real-world images with vanishing point annotations. 
Through a benchmark evaluation of four VPD algorithms - a baseline LSD, J-Linkage \cite{Toldo2008RobustMS}, NeurVPS \cite{zhou2019neurvps}, and GPVPD \cite{lin2022deep} against R-VPD,  our quantitative results show that R-VPD outperforms statistically significantly all four approaches on RPVP-Synthetic (Table \ref{tab:sr_rpvp_si}), and on RPVP-Real (Table \ref{tab:sr_rpvp_ri}; excluding NeurVPS-tmm17).  For fair comparison with NeurVPS-tmm17 we use the RPVP-Real Exclusive dataset, where R-VPD performs better  than this SOTA deep learning method (Table \ref{tab:sr_rpvp_ri_ex}) but not statistically significantly so, thus ``on par" on this real-world RP image set. 
Finally, when extending to images with and without RPs (for which R-VPD is not designed for), R-VPD performs second to the best 
with a statistically insignificant difference (with {\em p-value} =0.001 used as the threshold, Table \ref{tab:sr_tmm17}). 

 R-VPD has limitations on RP images with strong perspective and on non-RP images. 
In this work we focused on natural scenes with a single vanishing point (Figure~\ref{fig:rp_rectify}), not Manhattan-world scenes with multiple vanishing points such as those found in indoor or urban environments. We aim to extend current R-VPD to go beyond images with single VPs and to be more robust on general indoor/outdoor scenes.


\vspace{0.2cm}

\noindent \textbf{Acknowledgments.} This research is funded in part by NSF Award \#1909315 and in part by NSF Award \#1248076. We thank all the PSU LPAC members, especially Keaton Yukio Kraiger, for their valuable input.

\newpage
{\small
\bibliographystyle{ieee_fullname}
\bibliography{egbib}
}

\clearpage
\newpage

\appendix

\noindent{\Large\bf Appendix}

\input{appendix}

\end{document}

%% file: appendix.tex
\crefname{section}{Sec.}{Secs.}
\Crefname{section}{Section}{Sections}
\Crefname{table}{Table}{Tables}
\crefname{table}{Tab.}{Tabs.}



\definecolor{mygreen}{HTML}{00FF00}




\section{Scores for Forward Feature Selection}

As part of the GRASP algorithm discussed in Section~\ref{sec:VP_algo} for simultaneous discovery of visual words and RP instances, we perform forward feature selection to determine what new features to propose to add to the set of features in an existing visual word.  This process of feature expansion is carried out with an eye towards adding features that will provide good support for fitting implicit lines. This suitability is measured by a combination of three score functions.

\subsection{Linearity Score:}
This function takes at least 3 SIFT keypoints as input. We fit a line to the \texttt{(x,y)} locations of these keypoints using OpenCV’s \texttt{fitLine} function. The linearity score is then computed as the average perpendicular distance from each keypoint's \texttt{(x,y)} location to the fitted line.  The score thus represents how colinear the keypoint locations are. 

\subsection{Angle Score:}
The function takes an ordered list of keypoints that form a progression along a line and calculates the angular difference for consecutive triplets \texttt{(A, B,} and \texttt{C)} using the absolute difference between the absolute differences in angles from \texttt{A} to \texttt{B} and from \texttt{B} to \texttt{C}, which effectively measures the change in direction between these segments:
\[
\left| \left| A.angle - B.angle \right| - \left| B.angle - C.angle \right| \right|
\]

The accumulated angular differences are averaged over the number of triplet comparisons made, providing a normalized score that reflects the average rate of angular change per segment.

\subsection{Scale Score:}
Similar to the Angle Score, the Scale Score takes an ordered linear progression of keypoints and examines each triplet of consecutive keypoints \texttt{(A, B,} and \texttt{C)} to compute the absolute difference between the relative scale changes from A to B and from B to C. This is accomplished by determining the ratios of sizes between successive keypoints and comparing these ratios, thereby measuring how the scale factor evolves from one keypoint to the next:

\[
\left| \frac{A.size}{B.size}  - \frac{B.size}{C.size} \right|
\]  

These differences are accumulated and then normalized by the number of triplets assessed, yielding an average value that quantitatively describes the overall scale variability across the sequence of keypoints. A lower average difference indicates a higher consistency in scale among adjacent triplets of keypoints.

\subsection{Note about Units:}
The units of the Linearity Score are in pixels, the Angle Score is in radians, and the Scale Score is unitless. The minimum value for all three scores is 0. However, since we sum the scores, it is possible that the maximum value can exceed 1. The term $N^2$ in the ensemble score $S_C =S_L * e^{(S_A +S_S)}/N^2$ further normalizes the score by the number of features squared, potentially reducing the influence of large numbers of closely packed features.

\section{Datasets}
\label{supsec:dataset}
In this section, we provide more details about scene creation using Blender for RPVP-Synthetic (Recurring Pattern-based Vanishing Point Dataset with Synthetic Images) and labeling of RPVP-Real (Recurring Pattern-based Vanishing Point Dataset with Real-world Images) using the LabelMe tool.

\subsection{RPVP-Synthetic Image Dataset}
\begin{figure*}[!h]
\centering
\begin{subfigure}[t]{0.32\linewidth}
    \frame{\includegraphics[height = 2cm, width = \textwidth]{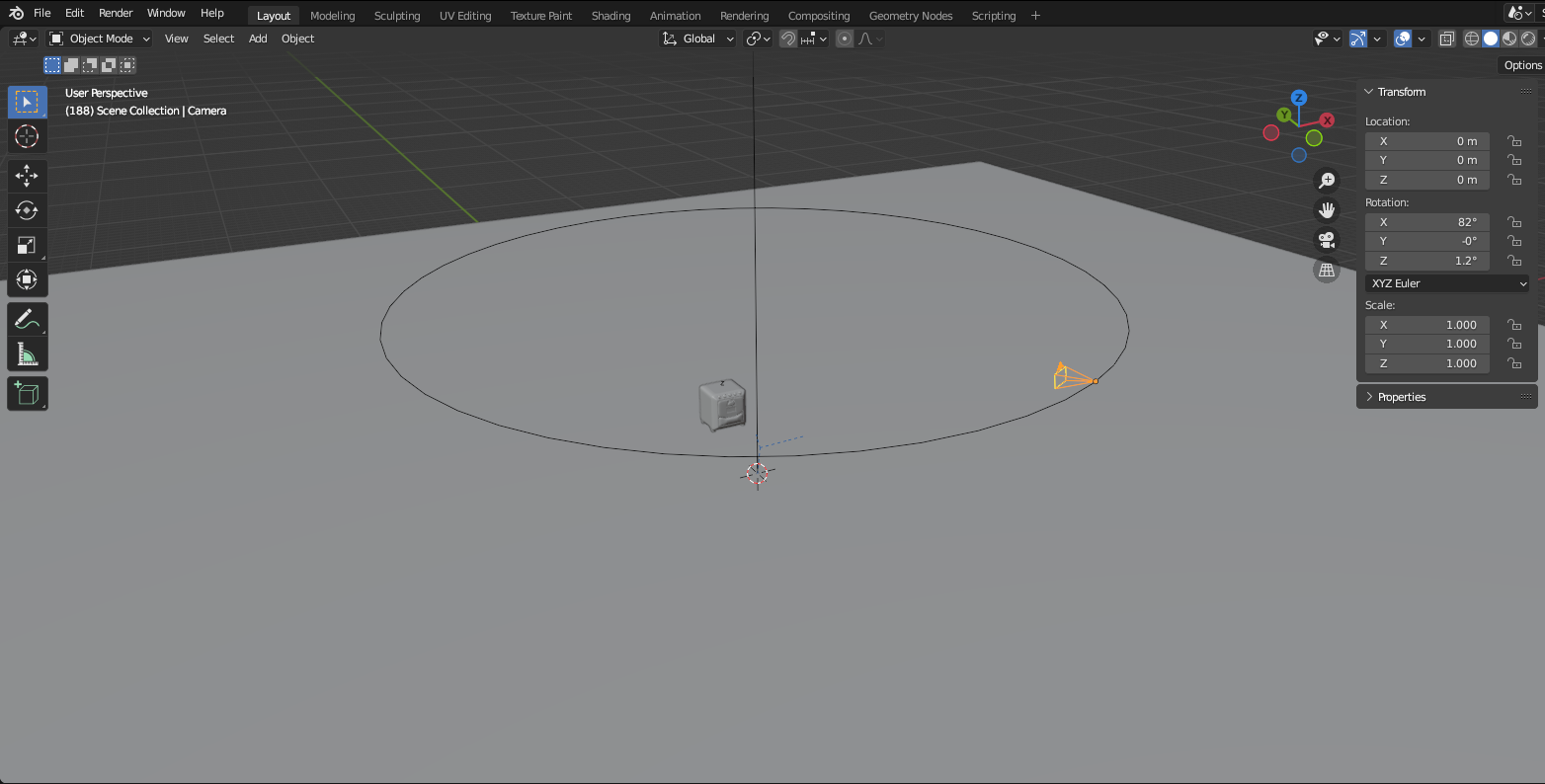}}
    \caption{Solid View}
    \label{fig:blend1}
\end{subfigure}\hfill
\begin{subfigure}[t]{0.32\linewidth}
    \frame{\includegraphics[height = 2cm, width = \textwidth]{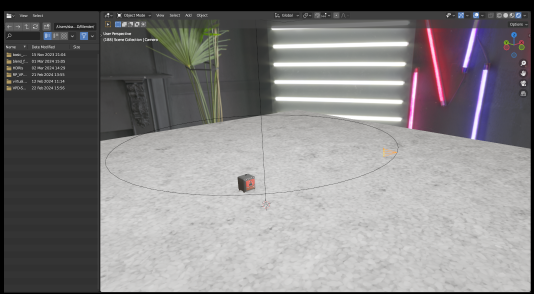}}
    \caption{Material View}
    \label{fig:blend2}
\end{subfigure}\hfill
\begin{subfigure}[t]{0.32\linewidth}
    \frame{\includegraphics[height = 2cm, width = \textwidth]{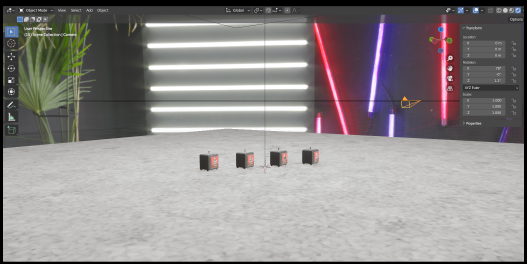}}
    \caption{Rendered View}
    \label{fig:blend3}
\end{subfigure}\hfill
\caption{The images in this figure represent stages of scene creation in Blender. (a) shows a screenshot of the \textit{Solid View} in Blender used to create the backbone of the scene. (b) shows the screenshot of the \textit{Material View} in Blender used to provide texture and backgrounds to the scene. And finally, (c) shows the screenshot of the final scene to be rendered.}  
\label{fig:blender_pipeline}
\end{figure*}

RPVP-Synthetic is a dataset of synthetic images created using an open-source 3D computer graphics software tool called \textit{Blender} \cite{Blender}. Blender allows one to create realistic 3D scenes with support for a 3D pipeline that includes modeling, animation, rendering, compositing, and motion tracking. For the RPVP-Synthetic dataset, we create multiple realistic 3D scenes with various backgrounds, surfaces, and objects. We then allow a camera to move along a circular trajectory and render the frames. Figure~\ref{fig:blender_pipeline} represents screenshots taken from Blender during different stages of creating a 3D scene.

For simplicity, we can divide the scene creation into three different steps. Step 1 (shown in Figure~\ref{fig:blend1}) is the initial step of creating the 3D scene in \textit{Solid View}. In this step, we create the backbone of the 3D scene that involves creating planes and objects, placing the camera, and providing it with a trajectory to move. It is important to note that the location of every object within the scene with respect to the default \textit{World Origin} of Blender is known. In step 2 (shown in Figure~\ref{fig:blend2}) we use \textit{Material View} to create visuals for the 3D scene. This step includes but is not limited to, adding backgrounds, providing objects a material view, and adding surface textures. Finally, in step 3 (shown in Figure~\ref{fig:blend3}) the camera is allowed to move in a prescribed trajectory and render images.

\begin{figure*}[!h]
\centering
\begin{subfigure}[t]{0.32\linewidth}
    \frame{\includegraphics[width = \textwidth]{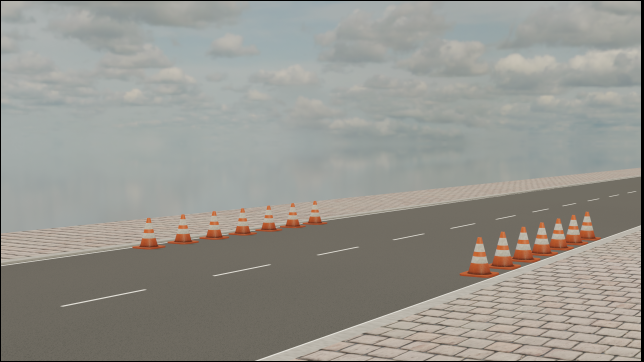}}\\[3pt]
    \frame{\includegraphics[width = \textwidth]{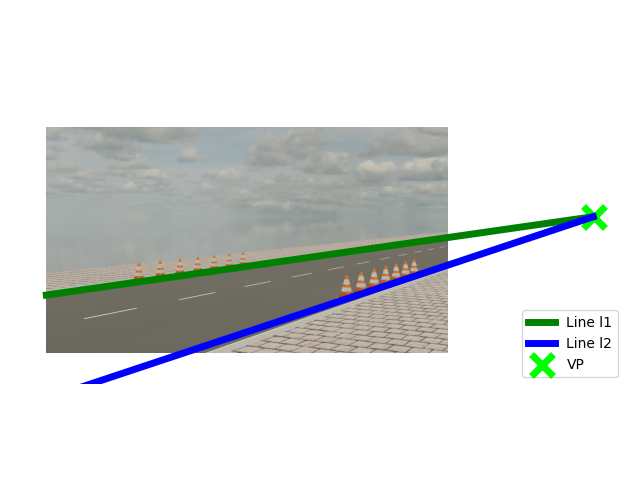}}
    \label{fig:blend_vp_1}
\end{subfigure}\hfill
\begin{subfigure}[t]{0.32\linewidth}
    \frame{\includegraphics[width = \textwidth]{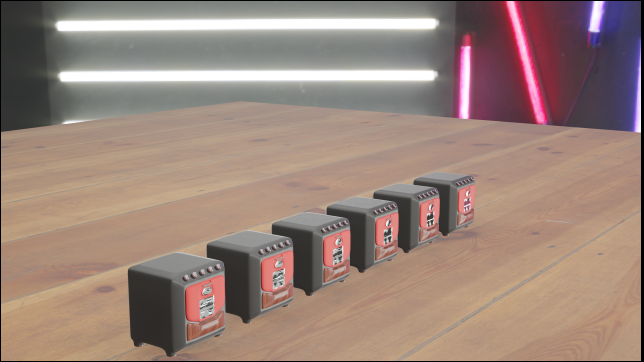}}\\[3pt]
    \frame{\includegraphics[width = \textwidth]{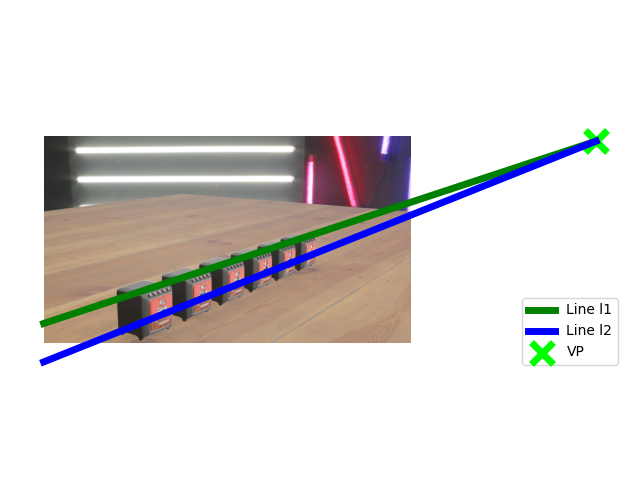}}
    \label{fig:blend_vp_1}
\end{subfigure}\hfill
\begin{subfigure}[t]{0.32\linewidth}
    \frame{\includegraphics[width = \textwidth]{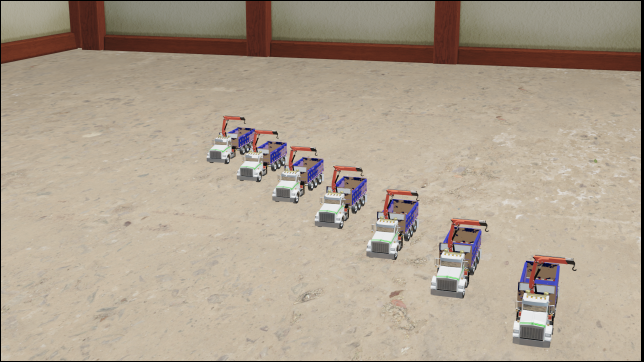}}\\[3pt]
    \frame{\includegraphics[width = \textwidth]{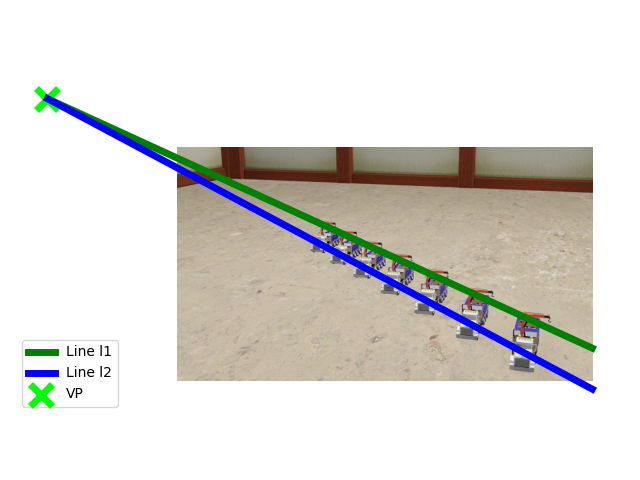}}
    \label{fig:blend_vp_1}
\end{subfigure}\hfill
\caption{Images in Row 1 represent sample frames ($1920\times1080$) rendered from Blender. Row 2 represents sample VP annotated images where lines \textcolor{ForestGreen}{$l1$} and \textcolor{blue}{$l2$} represent the projections of 3D scene parallel lines and the vanishing point (\textcolor{green}{$\times$}).}
\label{fig:blender_vp}
\end{figure*}

\subsubsection{Arranging Recurring Patterns to form Perspective View (Fig~\ref{fig:blender_vp}):} In the constructed 3D scene, an object is meticulously duplicated and arranged such that the arrangement appears to converge towards a single vanishing point.
Formally, the positioning of the objects is governed by the equation of a line in three-dimensional space, represented as $L(t) = O + tD$ where, $L(t)$ denotes the position of an object along the line at a parameter $t$, $O$ is the origin point in 3D space from which the objects are arranged, $D$ is the direction vector pointing towards the vanishing point, $t$ is a scalar that determines the position of each object along the line. The objects are placed such that their distances from the observer increase incrementally, causing them to appear smaller and smaller due to the perspective effect. This effect is quantitatively described by the perspective projection formula: $P(x, y, z) = (\frac{f \cdot x}{z}, \frac{f \cdot y}{z})$, where $P(x,y,z)$ represents the projection of a 3D point $(x,y,z)$ onto a 2D plane, $f$ is the focal length of the camera.

It is also worth mentioning that Blender does not follow the conventional camera coordinate system used commonly in computer vision. Blender uses a right-handed coordinate system where the Z-axis points upwards, the Y-axis points forward, and the X-axis points to the right. In addition, Blender does not directly provide rotations and locations of the scene objects with respect to the camera coordinate system. Therefore, extrinsic parameters were transformed from the Blender coordinate system to a computer vision-friendly coordinate system by systematically applying matrix transformations. 

\begin{figure*}[!h]
\centering
\begin{subfigure}[t]{0.32\linewidth}
    \frame{\includegraphics[width = \textwidth]{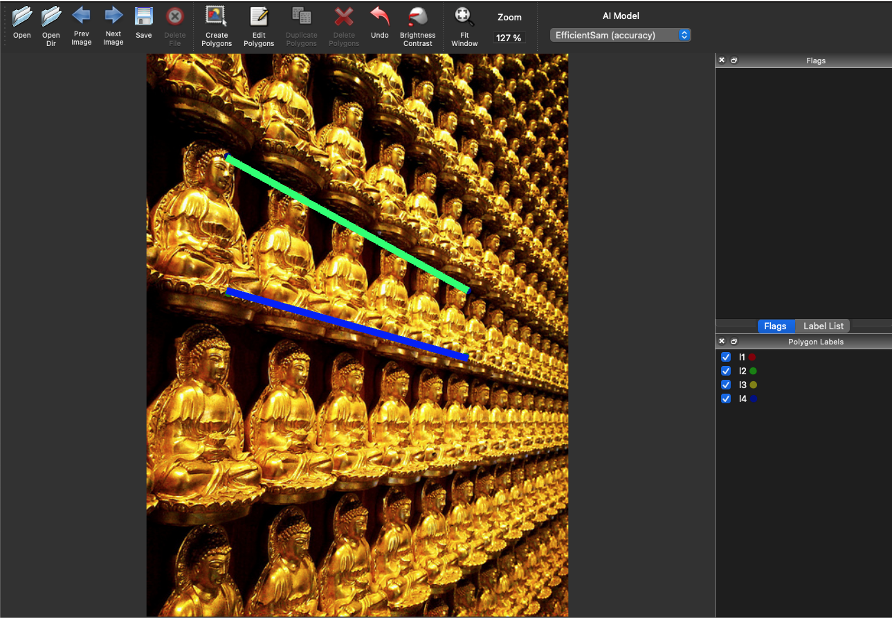}}\\[3pt]
    \frame{\includegraphics[width = \textwidth]{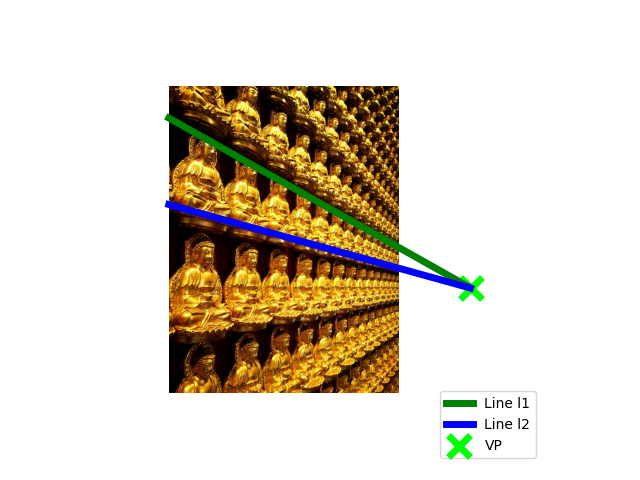}}
    \label{fig:labelme_1}
\end{subfigure}\hfill
\begin{subfigure}[t]{0.32\linewidth}
    \frame{\includegraphics[width = \textwidth]{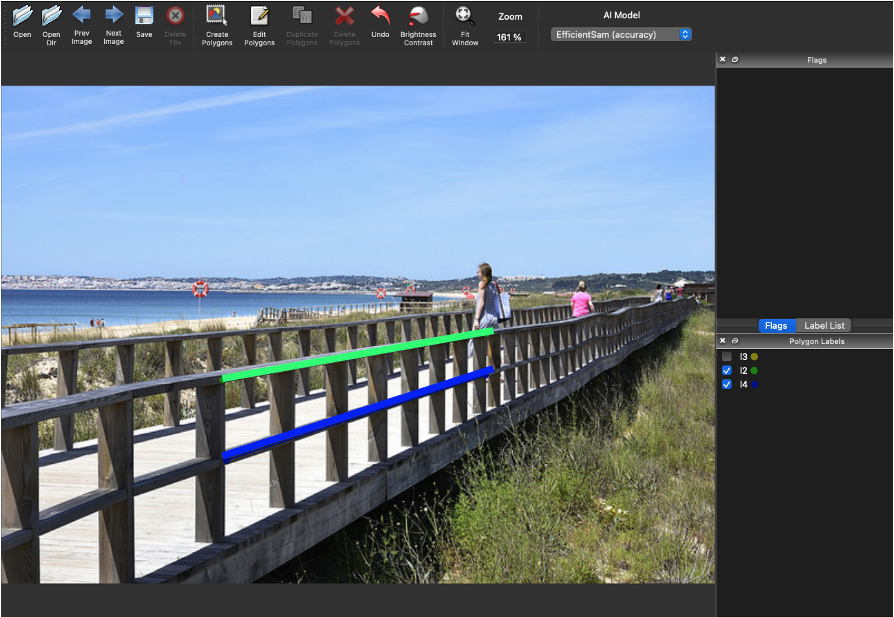}}\\[3pt]
    \frame{\includegraphics[width = \textwidth]{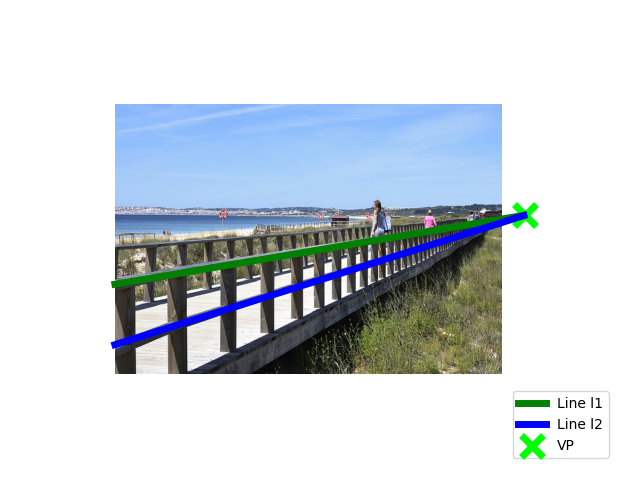}}
    \label{fig:labelme_2}
\end{subfigure}\hfill
\begin{subfigure}[t]{0.32\linewidth}
    \frame{\includegraphics[width = \textwidth]{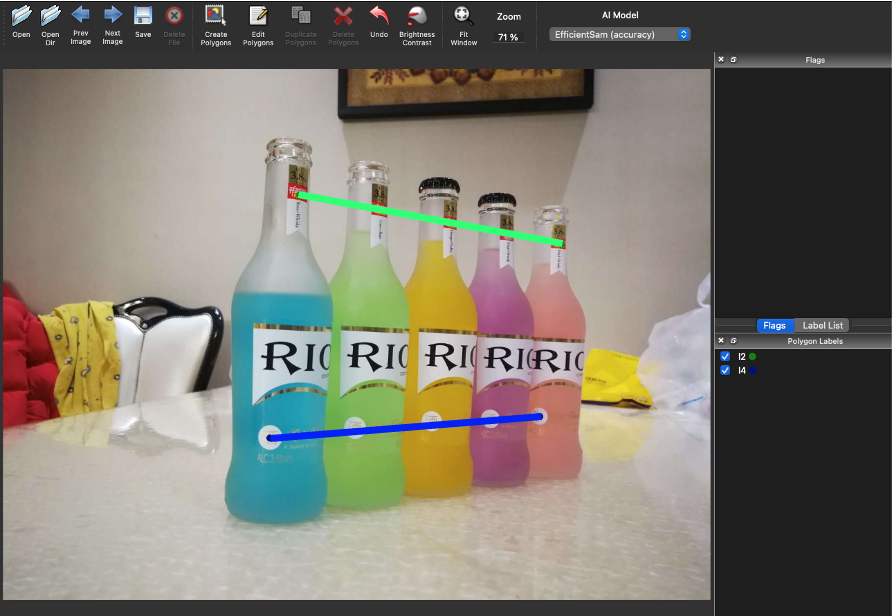}}\\[3pt]
    \frame{\includegraphics[width = \textwidth]{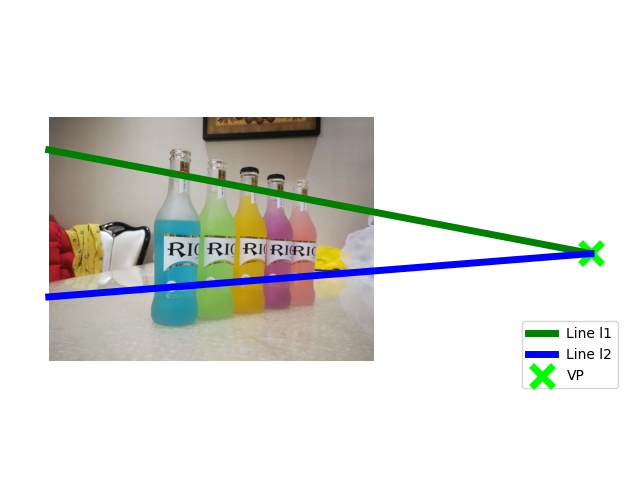}}
    \label{fig:labelme_4}
\end{subfigure}\hfill
\caption{Images in Row 1 represent screenshots from LabelMe. Row 2 represents sample VP annotated images where lines \textcolor{ForestGreen}{$l1$} and \textcolor{blue}{$l2$} represent the annotated lines and the vanishing point (\textcolor{green}{$\times$}).}
\label{fig:labelme}
\end{figure*}

\subsubsection{Locating Ground Truth Vanishing Point:} In a theoretical context, the ground truth VP of parallel scene lines with orientation $D'$ will be the intersection of a viewing ray $kD$ from the camera origin, intersecting with the image plane $Z=f$, where $D=RD'$ is the 3D line orientation $D'$ described in the camera coordinate system with a relative rotation $R$. Specifically, consider a set of parallel lines that pass through the recurring patterns with an orientation $D = (D_x, D_y, D_z)$ within the camera's coordinate system, the vanishing point for a camera with a focal length $f$ can be concisely represented by the coordinates $(\frac{f \cdot D_x}{D_z}, \frac{f \cdot D_y}{D_z})$.

On the other hand, an empirical approach can also be used to establish the ground truth vanishing point based on the projection of specific geometric entities. Consider four distinct points ($P_1, P_2, P_3, P_4$) in 3D space, which serve as the endpoints of two parallel scene lines that recede into the distance. Each of these selected 3D points is projected onto the 2D image plane yielding four corresponding 2D points ($P'_1, P'_2, P'_3, P'_4$). Two lines are constructed in the 2D plane, one connecting $P'_1$ and $P'_2$, and the other connecting $P'_3$ and $P'_4$. These (implicit) lines represent the perspective projections of the original 3D parallel lines. The point of intersection of these two lines in the 2D plane is computed, locating the vanishing point in the image for the pair of original parallel lines in 3D space. 

While both the theoretical and empirical approaches yield the same vanishing point, the latter is much easier to analyze and visualize. Figure~\ref{fig:blender_vp} shows some example images of the original rendered frames from different scenes (Row 1 of Figure~\ref{fig:blender_vp}) and the corresponding ground truth VP calculated as the intersection of the two lines (Row 2 of Figure~\ref{fig:blender_vp}).  

\subsubsection{RPVP-Synthetic Dataset Files:} As a part of the publication, we intend to provide the following files: 1) rendered frames ($1920\times1080$), 2) ground truth VP for each frame as a text file, and 3) camera parameters for each frame used to obtain these projections also as text files. This dataset consists of 3,200 images extracted from 16 different scenes with each frame contributing 200 images.

\subsection{RPVP-Real World Image Dataset}
RPVP-Real is a dataset of real-world images collected from publicly available images online. These images include images from Google, Flickr, self-photographed images, etc. There are a total of 1400 images that are manually labeled by researchers using an open-source graphical image annotation tool called \textit{LabelMe} \cite{LabelMe}. The LabelMe tool allows users to annotate images with polygons, rectangles, circles, lines, etc. Similar to RPVP-Synthetic, the vanishing point for these images is calculated by finding the intersection of lines that pass through the recurring patterns. But in this case, the two lines are drawn manually on the image over the recurring patterns. Figure~\ref{fig:labelme} represents screenshots of the LabelMe with the annotated lines (Row 1) and the corresponding images (Row 2) with the calculated ground truth vanishing point. 

\subsubsection{RPVP-Real Dataset Files:} As a part of the publication, we intend to provide the following files: 1) images, and 2) ground truth VP for each frame as a text file. This dataset consists of a total of 1,400 images.

\begin{figure}[h!]
\centering
\begin{subfigure}[t]{0.49\linewidth}
    \centering
    \includegraphics[width=\linewidth]{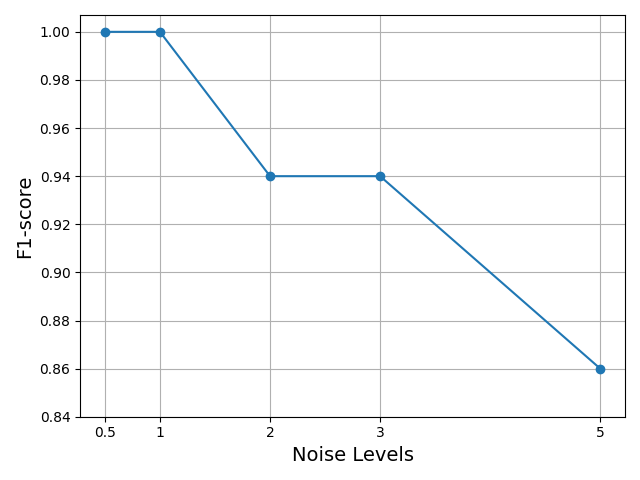}
    \caption{@ Angle Threshold of $5^\circ$.}
\end{subfigure}\hfill
\begin{subfigure}[t]{0.49\linewidth}
\centering
    \includegraphics[width=\linewidth]{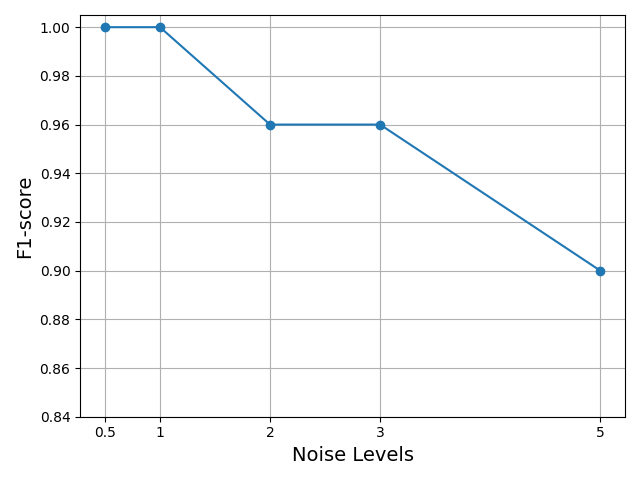}
    \caption{@ Angle Threshold of $10^\circ$.}
\end{subfigure}\hfill
\caption{Robustness of our vanishing point detection system is evaluated using a stress test. We methodically vary the feature densities by resizing images to simulate different levels of feature availability and manipulate spatial distributions by introducing Gaussian noise at various standard deviations ([0.5, 1, 2, 3, 5]) to the location of the feature points. F1-scores are computed at each noise level. Figure~\ref{fig:stress_test} shows some sample images.}
\label{fig:f1_score}
\end{figure}


\section{Processing Time Analysis}
The processing times of the four algorithms were evaluated on real-world and synthetic datasets, with a focus on median times for a robust central measure due to result variability.

NeurVPS and GPVPD were the fastest on real-world images, with median times of 0.86 and 0.48 seconds, respectively, while J-Link and R-VPD took longer, with medians of 18.67 and 10.73 seconds. On the synthetic dataset, NeurVPS and GPVPD again showed quick processing (0.86 and 0.58 seconds), while J-Link and R-VPD had longer medians of 0.75 and 1.31 seconds. The lack of straight-line segments reduced J-Link’s processing time, though this resulted in poorer VPD accuracy (Figure~\ref{fig:sr_rpvp_si}).

\vspace{0.1cm}

\noindent \textbf{Key Observations:}
\begin{enumerate}
    \item Impact of Optimization and GPU Utilization: NeurVPS and GPVPD benefit from GPU usage and optimized deep learning libraries, resulting in faster processing, whereas R-VPD and J-Link, relying on CPUs and less optimization, have longer times.

    \item Trade-off Between Time and Space: NeurVPS and GPVPD’s speed advantage comes with high memory demands (~400 MB per model), which can be limiting in memory-constrained environments where lightweight models might be preferred despite slower processing times.
\end{enumerate}

\vspace{0.1cm}

The comparison highlights each approach's trade-offs: NeurVPS and GPVPD are faster due to GPU and optimization but require substantial memory, while J-Link and R-VPD, though slower, offer more consistency in CPU-only settings. Median times provide a fair representation of each algorithm's typical performance.

\section{Stress Test}
\begin{figure*}
\centering
\includegraphics[width=\linewidth]{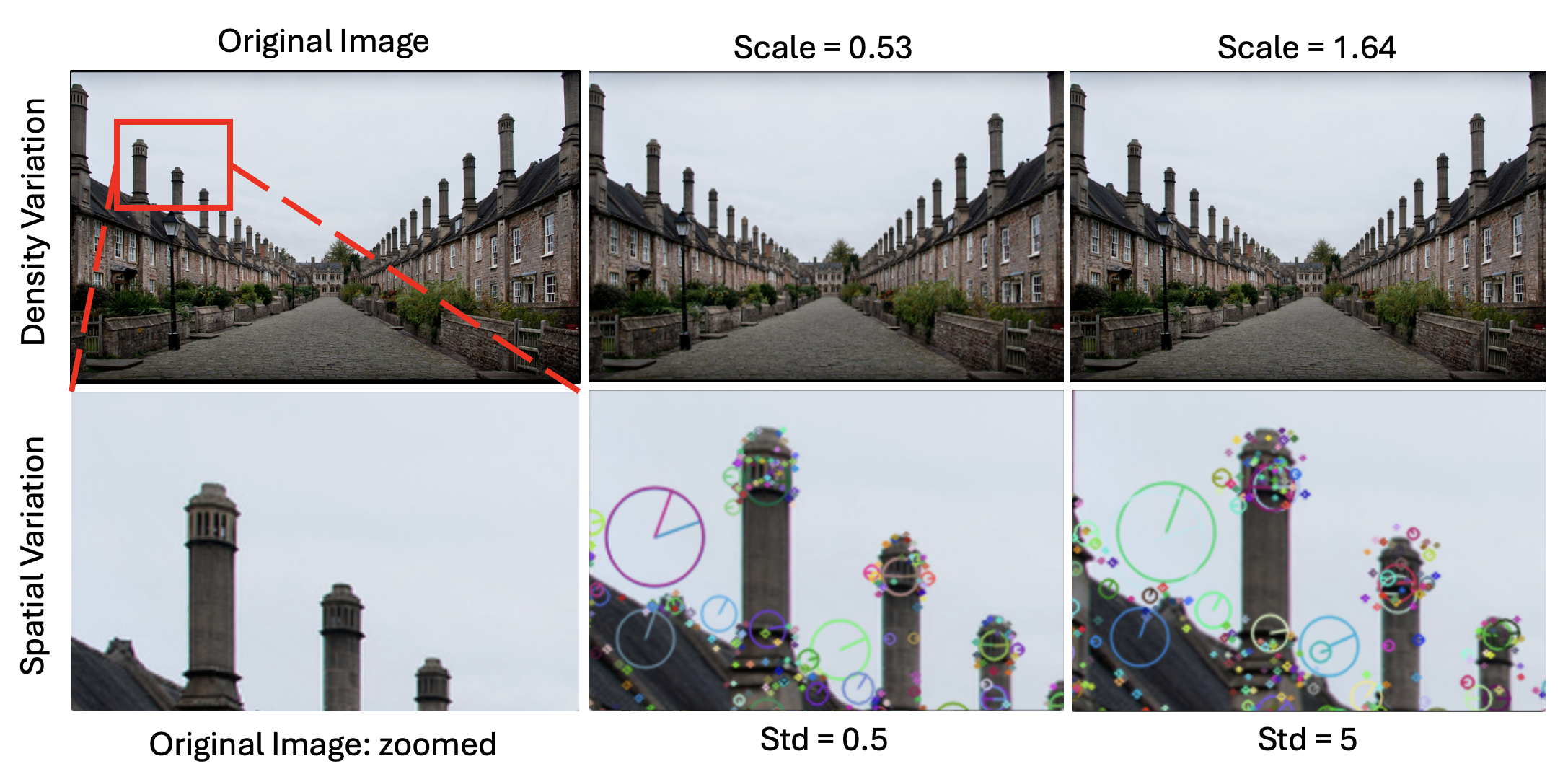}
\caption{Top row: original image and scaled images that introduce density variation of SIFT features. Bottom row: zoomed-in image patches representing SIFT features with noise added to their location that perturbs the spatial distribution of the features.}
\label{fig:stress_test}
\end{figure*}

To emphasize the robustness of our vanishing point detection system, we conducted a controlled stress test \cite{liu2013grasp}. This test methodically varies feature densities by resizing images to simulate different levels of feature availability and manipulates spatial distributions by introducing Gaussian noise at various standard deviations ([0.5, 1, 2, 3, 5]) to the location of the feature points. We calculated the \textit{F1}-score for each noise level at angle thresholds $5^\circ$ and $10^\circ$. The consistency between precision and recall at each threshold simplified our analysis, allowing us to use \textit{F1}-scores to evaluate performance directly. Figure~\ref{fig:f1_score} represents \textit{F1}-scores calculated at the aforementioned noise levels at angle thresholds $5^\circ$ and $10^\circ$. Figure~\ref{fig:stress_test}
represents the variations in the input image. The top row represents scaled input images that vary the feature density in the input image and the bottom row represents images where noise is added to the location of the SIFT features to manipulate spatial distribution. It can be seen that the SIFT features in the rightmost image (bottom row) with $std = 5$ is more spread out compared to the middle image (bottom row) with $std = 0.5$. These robust results validate the effectiveness of our approach when faced with significant alterations in feature density and spatial arrangement, confirming its applicability across diverse operational environments.

\section{More Qualitative Examples}
\begin{figure*}[h!] 
\centering
\begin{subfigure}[t]{0.19\linewidth}\centering
\caption{GT}
    \frame{\includegraphics[height = 0.90\textwidth, width =1\textwidth]{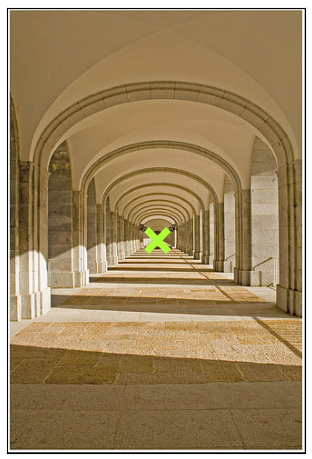}}
    \frame{\includegraphics[height = 0.90\textwidth, width =1\textwidth]{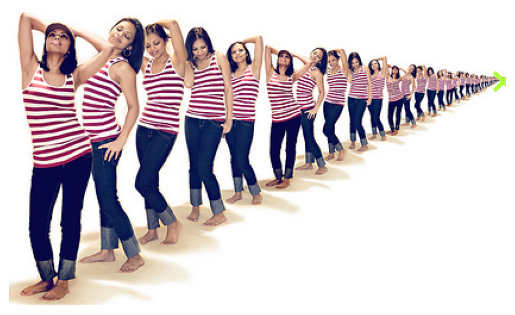}} 
\end{subfigure}
\begin{subfigure}[t]{0.19\linewidth}\centering
\caption{SIFT}
    \frame{\includegraphics[height = 0.90\textwidth, width =1\textwidth]{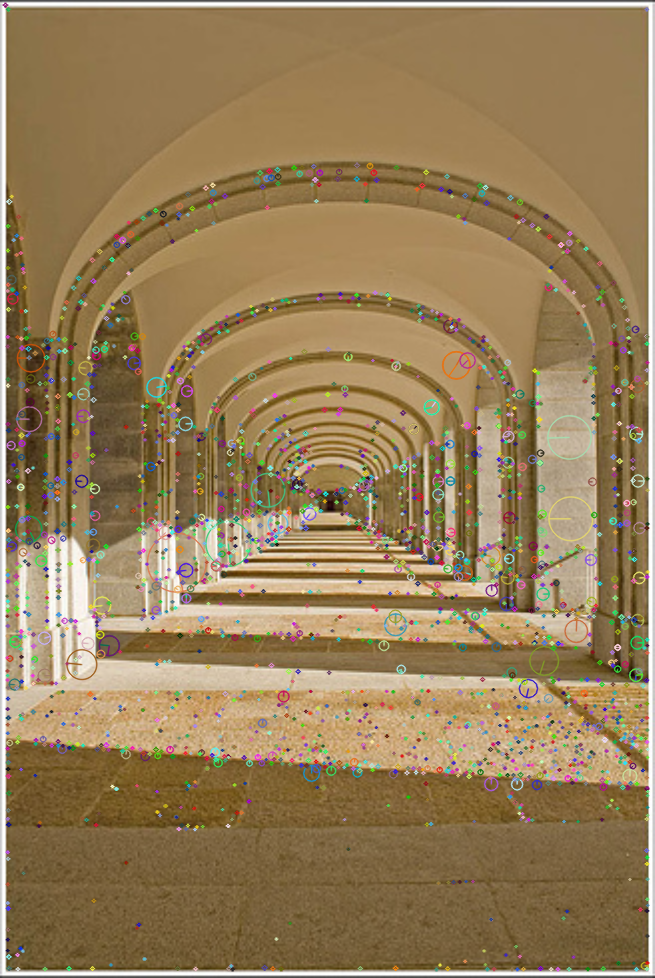}}
    \frame{\includegraphics[height = 0.90\textwidth, width =1\textwidth]{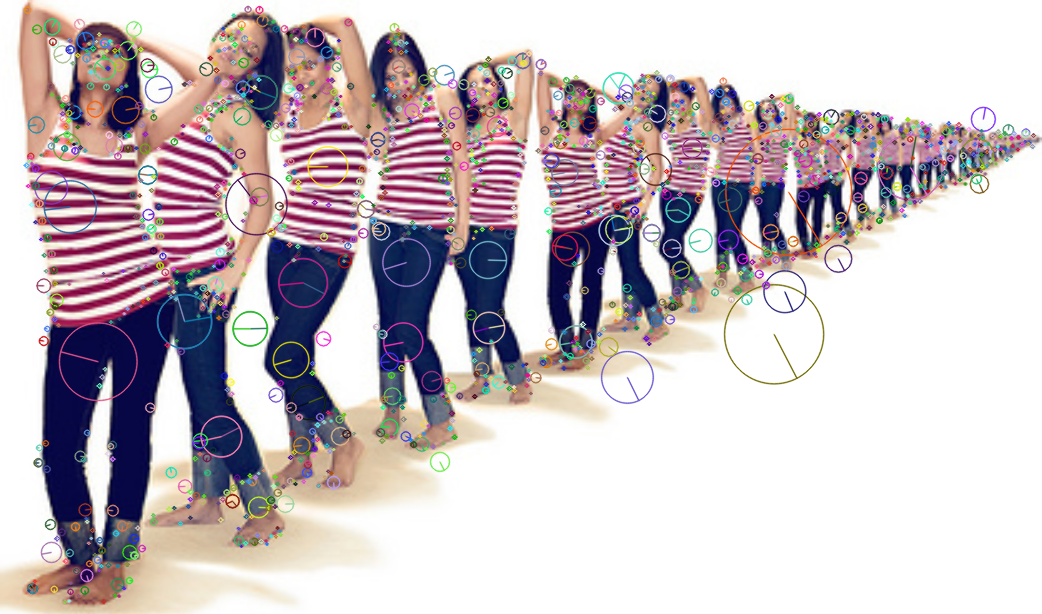}}
\end{subfigure}
\begin{subfigure}[t]{0.19\linewidth}\centering
\caption{Line Fit}
    \frame{\includegraphics[height = 0.90\textwidth, width =1\textwidth]{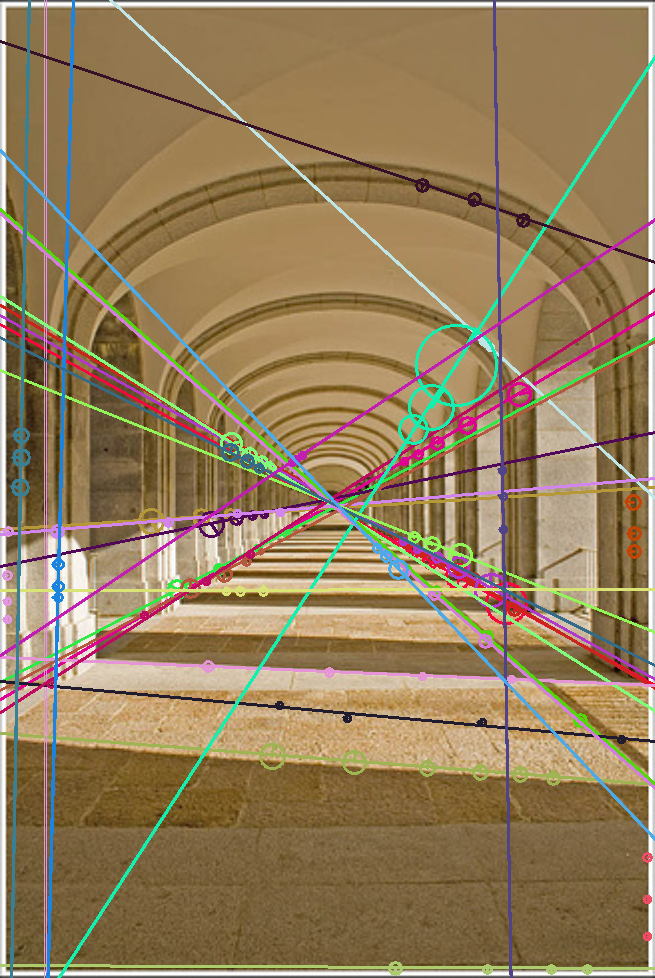}}
    \frame{\includegraphics[height = 0.90\textwidth, width =1\textwidth]{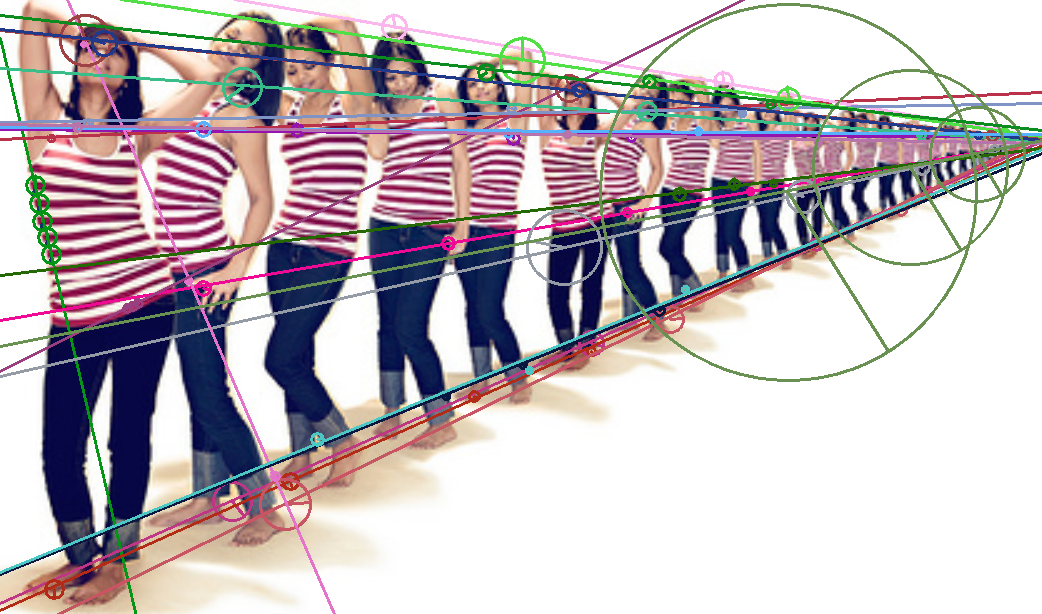}}
\end{subfigure}
\begin{subfigure}[t]{0.19\linewidth}\centering
\caption{Inliers}
    \frame{\includegraphics[height = 0.90\textwidth, width =1\textwidth]{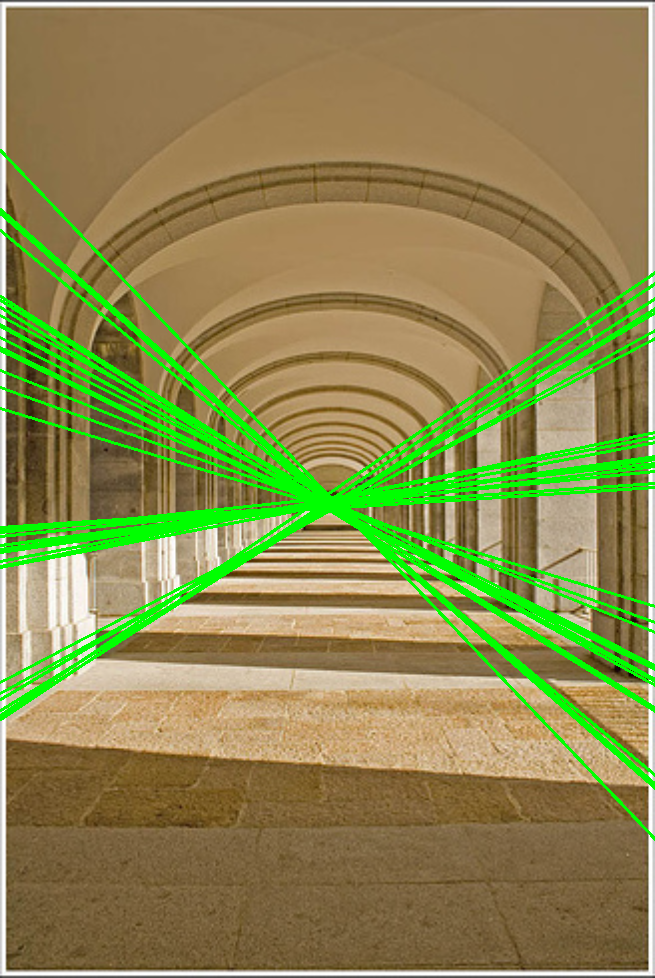}}
    \frame{\includegraphics[height = 0.90\textwidth, width =1\textwidth]{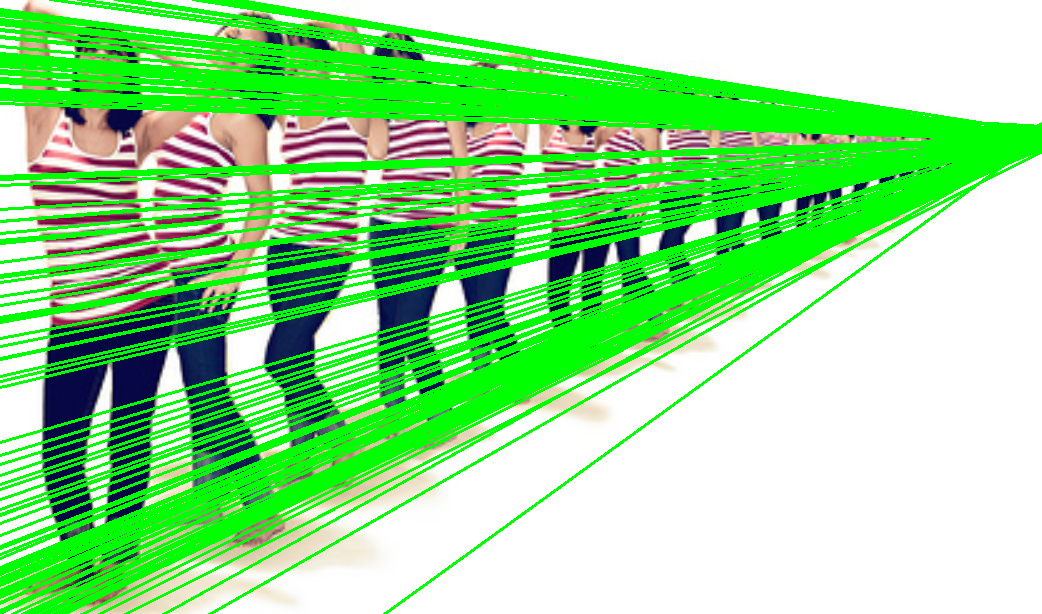}} 
\end{subfigure}
\begin{subfigure}[t]{0.19\linewidth}\centering
\caption{Detected}
    \frame{\includegraphics[height = 0.90\textwidth, width =1\textwidth]{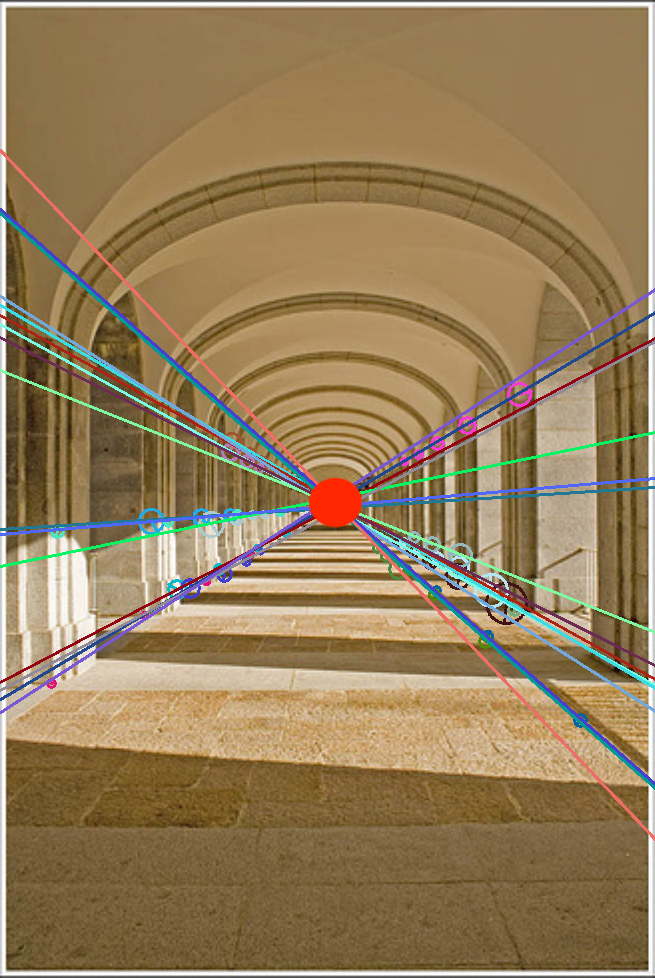}}
    \frame{\includegraphics[height = 0.90\textwidth, width =1\textwidth]{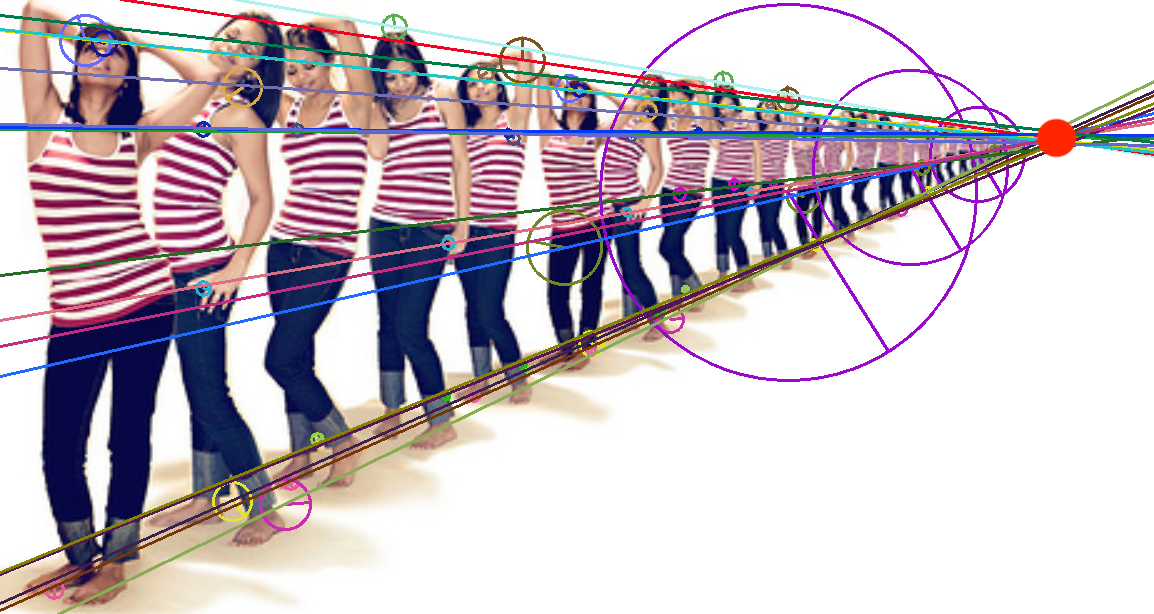}} 
\end{subfigure}
\caption{Sample results from our R-VPD algorithm with images from intermediate stages.
\textbf{(a)} Input image with ground truth (shown as \textcolor{green}{$\times$}), 
\textbf{(b)} SIFT features extracted, 
\textbf{(c)} lines fitted to SIFT feature groups,
\textbf{(c)} inlier set from weighted RANSAC,
\textbf{(e)} VP detected from our method (\textcolor{red}{Red} circle).}
\label{fig:sup_rpvpd}
\end{figure*}

The following pages provide a wide range of sample results from our experiments. Figure~\ref{fig:sup_rpvpd} represents images from different stages of our R-VPD algorithm. The images are shown in sequence from left to right starting with the original image with annotated ground truth. Column 2 represents images with SIFT features. Column 3 represents images with implicit lines fit on selected feature correspondences. Column 4 shows images with the inlier set obtained using weighted RANSAC. Finally, column 5 shows the detected vanishing point. 

\begin{figure*}[h!]
\centering
\includegraphics[width = \linewidth]{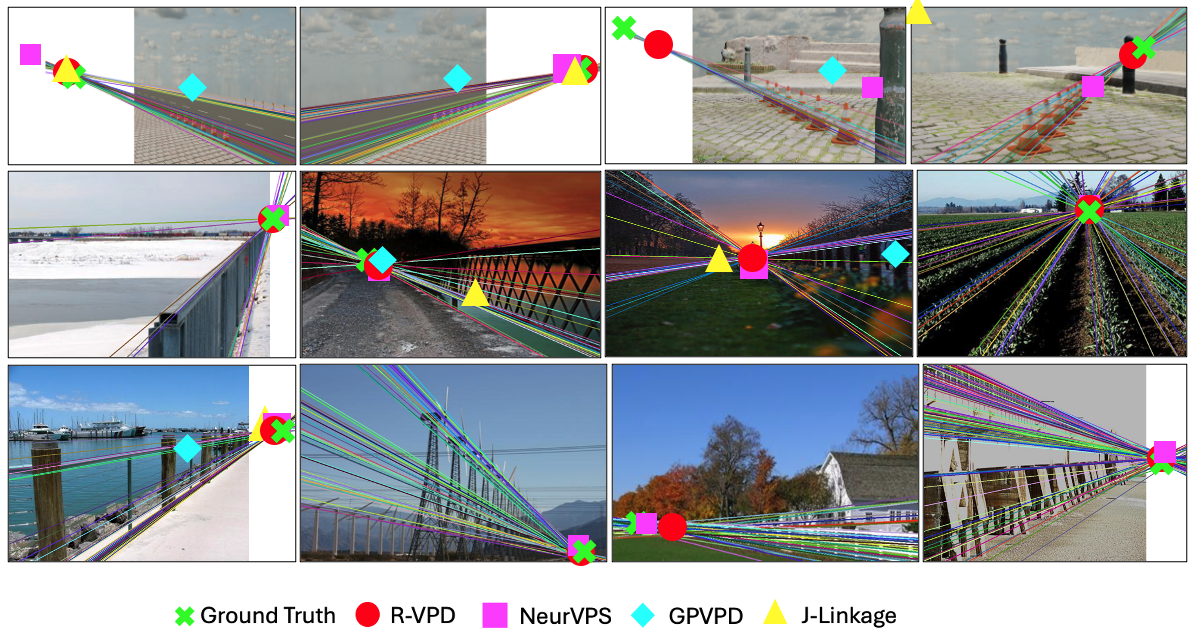}
\caption{Sample results for VP detection on RPVP-Synthetic (Row 1), RPVP-Real (Row 2), and TMM17 (Row 3) using our method and a comparison with other approaches.}
\label{fig:sup_vp_ex}
\end{figure*}

\begin{figure*}
\centering
\includegraphics[width = \linewidth]{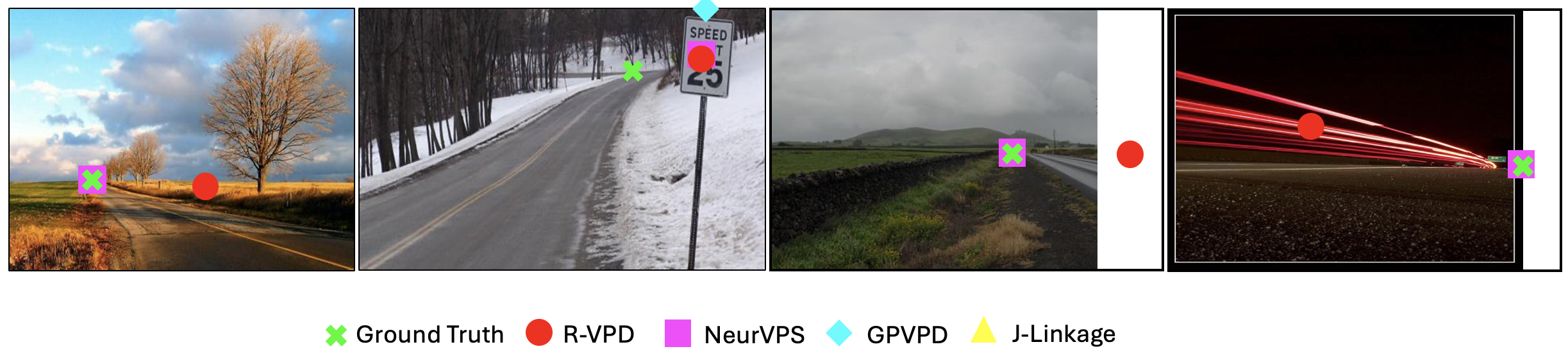}
\caption{Failure cases of our method. 
The images are all from the TMM17-test dataset (Figure \ref{fig:dataset}). Failures are mostly attributed to the absence of RPs as well as straight-line segments in the image.}
\label{fig:sup_failure}
\end{figure*}

Figure~\ref{fig:sup_vp_ex} represents more examples from the RPVP-Synthetic dataset that compare the outputs of R-VPD with the outputs of different methods. Similarly, Figure~\ref{fig:sup_vp_ex} and Figure~\ref{fig:sup_vp_ex} show more examples comparing the outputs of R-VPD with the outputs of different methods on RPVP-Real and TMM17 datasets, respectively. Finally, Figure~\ref{fig:sup_failure} illustrates some failure cases. A more robust recurring pattern detection algorithm operating across different scales would greatly improve our R-VPD approach on these challenging images.

